\documentclass[12pt,a4paper]{article}
\usepackage{graphicx}
\usepackage{amsmath}
\usepackage{geometry}
\usepackage{booktabs}
\usepackage{amssymb}
\usepackage{enumitem}
\usepackage{titlesec}
\usepackage[style=apa]{biblatex} 
\usepackage{hyperref}
\usepackage{float}
\usepackage[utf8]{inputenc}
\usepackage{pgfgantt}
\usepackage{tikz}
\usetikzlibrary{arrows.meta, positioning, fit}

\definecolor{hardwarecolor}{RGB}{138, 99, 210}
\definecolor{dataacqcolor}{RGB}{52, 168, 83}
\definecolor{preprocesscolor}{RGB}{251, 188, 4}
\definecolor{modelcolor}{RGB}{66, 133, 244}
\definecolor{integrationcolor}{RGB}{255, 105, 180}

\usepackage{subcaption} 
\title{\textbf{Robotic Arm-Based Spectral Sensing for Strawberry Positioning and Non-Destructive Sweetness Measurement}}
\author{Yi Yang\\ Postdoctoral Supervisor: Dr Mark Cardamis\\ Supervisor: Professor Wen Hu}
\date{May 2026}

\begin{document}

\maketitle
\begin{abstract}

Accurate assessment of fruit sweetness is essential for quality control in modern agriculture, yet conventional methods still rely on destructive sampling and are difficult to scale. This thesis presents a robotic arm-based spectral sensing system for automated strawberry detection, localization, approach, and non-destructive sweetness estimation.

The system integrates perception, calibration, and robotic control in a closed-loop pipeline. A YOLOv11s detector is adopted for real-time strawberry detection, while RGB--ToF calibration and mask-to-depth alignment are used to obtain geometrically consistent target localization. A custom eye-in-hand hand--eye calibration workflow is developed to estimate the rigid transform between \texttt{gripper\_link} and \texttt{cam\_front}, enabling reliable transformation of fruit targets into the robot base frame. Based on these estimates, the robot executes a waypoint-based search and an incremental closed-loop approach strategy to position the sensor at an optimal working distance for sweetness sensing.

Experimental results show strong end-to-end performance (88.10\% success over 42 trials), with robust detection (95.24\%) and fully successful approach execution once a target is detected (100\% conditional success). Hand--eye calibration comparisons indicate that although Andreff yields the smallest translation norm in single-run results, the Park method provides better cross-sample consistency and therefore more stable downstream robot behavior. The main residual failures are concentrated in the sensing stage, especially valid-region extraction for sweetness estimation under difficult depth/reflectance conditions.

Overall, this work demonstrates the feasibility of integrating RGB--ToF perception, robotic manipulation, and non-destructive sensing for practical strawberry quality assessment, and provides a scalable baseline for future integration of learning-based policies such as Vision--Language--Action (VLA) models.

\end{abstract}

\clearpage
\tableofcontents

\listoffigures
\listoftables

\section{Motivation}
Strawberries are a high-value horticultural crop, and their sweetness, commonly quantified by degrees Brix \parencite{brix}, is a key determinant of consumer preference and market value. Traditional methods for assessing sweetness rely on manual sampling and destructive testing, which are labor-intensive, time-consuming, and provide only limited coverage of large-scale production \parencite{smartref}.  Moreover, accurately identifying the positions of individual strawberries in complex plant environments is challenging, as visual occlusion and dense foliage can hinder effective inspection. As a result, only a small subset of fruits can be sampled, leading to inconsistent quality assessment across batches. This has created a demand for non-destructive, reliable techniques that can rapidly estimate sweetness while preserving the fruit.

In parallel, accurately detecting and localizing strawberries in real-world environments presents significant challenges. Strawberries often grow in dense clusters, partially occluded by leaves, stems, and surrounding fruits. Variations in lighting, background complexity, and fruit appearance further complicate reliable detection and segmentation.

Recent advances in spectral sensing and robotic manipulation provide an opportunity to address both challenges simultaneously. By mounting spectral and depth sensors on a robotic arm, it becomes possible to actively scan the environment, automatically detect strawberries, precisely localize them in 3D space, and position the sensor at an optimal distance for measurement. This enables a fully automated pipeline in which the system not only identifies the target fruit but also returns an estimate of its sweetness without any physical contact or damage.

The motivation for this project stems from the need to improve efficiency, accuracy, and scalability in strawberry quality assessment. Non-destructive methods allow for rapid evaluation without damaging the fruit, facilitating standardized grading across large operations. Furthermore, arm-mounted spectral sensing systems can be adapted to other fruit types, extending the impact of this approach beyond strawberries.

This project therefore seeks to develop a robotic arm-mounted spectral sensing platform that integrates hardware setup, controlled data acquisition, computer vision models, and system integration to achieve accurate strawberry positioning and non-destructive sweetness measurement. Unlike existing commercial solutions, which often rely on destructive sampling, limited manual inspection, or single-modality sensors, this system combines RGB imaging and ToF depth and amplitude sensing to enable precise, non-destructive, and high-throughput assessment of strawberry sweetness directly on the plant, offering enhanced scalability and accuracy for greenhouse operations.

\section{Literature Review}
\subsection{Non-Destructive Sweetness Measurement}

Previous research has explored non-contact methods like mmWave radar with machine learning \parencite{tavasoli2022}, but these often lack the necessary resolution for fruits as mmWave are focusing on distance and motion detection. In contrast, integrating NIR spectroscopy with depth information and artificial intelligence, as demonstrated in projects like “SweetFruit,”  \parencite{cardamis2025} has shown much greater potential for accurate, non-destructive sweetness prediction for apple.  The system achieved an RMSE of approximately 0.57°Brix, indicating high accuracy. Despite this success, the approach was limited to a single apple variety (Granny Smith), and its generalizability to other fruits, such as strawberries, remained unclear.
Additionally, other recent studies have validated that NIR spectroscopy is effective for assessing the sweetness of strawberries as well \parencite{wold2024}. Still, the unique surface and environmental sensitivity of strawberries present challenges that this project aims to address.

\subsubsection{Principle for Detecting Sugar Based on ToF}

\paragraph{NIR spectroscopy $\rightarrow$ sugar (SSC, $^{\circ}$Brix)}
Sugars (and water) contain O--H, C--H, and C--O bonds whose overtone and combination vibrations absorb in the near-infrared (NIR; approximately 700--2500~nm). The absorbance at characteristic bands correlates with soluble solids content (SSC), enabling non-destructive prediction of $^\circ$Brix via multivariate regression models such as PLSR or neural networks \cite{R1,R2}. Recent reviews and commodity-focused studies (e.g., strawberry and citrus) report robust SSC performance in both laboratory and online/inline implementations \cite{R1,R2}.

\subsubsection*{ToF in Two Senses}
\paragraph{(i) Time-of-flight depth cameras (3D)}
These sensors measure per-pixel distance/shape and often provide an intensity/reflectance channel. Although they do not directly sense sugar, 3D cues (pose, curvature, contact quality) help correct spectral geometry, stabilise NIR readings, and add \mbox{size/firmness} proxies for grading \cite{R3,R4}.

\paragraph{(ii) Time-resolved NIR spectroscopy}
In time-resolved NIR (TRS), picosecond pulses are injected and the photon time-of-flight distribution is recorded to estimate the absorption coefficient ($\mu_a$) and the reduced scattering coefficient ($\mu'_s$). These optical properties relate to tissue composition; studies have predicted SSC and acidity (e.g., grapefruit) and compared TRS with conventional continuous-wave NIR \cite{R5}.

\subsection{Strawberry localization detection and segmentation}
\subsubsection{Chessboard Calibration}
Work by \parencite{Jung2014ToFCalibration} addresses the calibration of Time-of-Flight (ToF) depth sensors paired with color cameras, aiming to improve RGB-D data accuracy. The authors employ a specially designed 2.5D calibration board, which allows robust detection of correspondences in both color and depth modalities. They propose a per-pixel calibration approach that corrects both the ray directions and range biases of ToF measurements, followed by pose optimization to align depth and color images. Experimental results demonstrate significant improvements over conventional calibration methods, yielding more precise 3D reconstructions. This method is particularly relevant for robotic manipulation tasks requiring accurate spatial perception. For strawberry-picking robots, precise RGB-D calibration with such a 2.5D board can enhance fruit detection, depth estimation, and grasp planning, especially in complex environments with occlusions and irregular foliage. Accurate sensor fusion enables more reliable manipulation, reducing the risk of collision and improving the efficiency and success rate of autonomous harvesting.

\subsubsection{Fruit Localization}
\parencite{yu2024tactile} provided important insights for precise fruit positioning. The Z-coordinate was obtained from a Time-of-Flight (ToF) sensor, while geometric calibration allowed the X and Y coordinates to be derived through a linear transformation from 2D pixel coordinates. This method is particularly suited for complex or confined spaces, improving target localization and manipulation accuracy. However, the approach has several limitations: the ToF sensor could potentially replace a ZED2 camera for depth acquisition; YOLOv4-based detection was slower and less accurate compared to more recent models; the computational load was relatively high; and there was a risk of accidental damage during physical manipulation.

\parencite{lee2024sonicboom} employed ultrasonic sensors for fruit detection, which allowed the system to operate independently of lighting conditions, making it suitable for dark environments. It achieved high localization accuracy. Nevertheless, the approach required physical contact to detect objects, making it unsuitable for non-contact applications and potentially limiting its use in delicate fruit handling.

\subsection{Robotic Arm Strawberry Search and Approach}
\subsubsection{Hand-eye calibration}

\parencite{MathWorks2024MovingCameraPose} shows how to perform and verify hand-eye calibration for a robot arm or manipulator equipped with a camera in the eye-in-hand configuration (also referred to as a moving camera configuration). In this setup, the camera is rigidly attached to the end-effector, and the objective is to estimate the transformation between the camera frame and the robot end-effector frame.

Mathematically, hand-eye calibration is commonly formulated as the well-known $AX = XB$ problem, where $A$ represents the motion of the robot end-effector between two poses, $B$ represents the corresponding motion of the camera, and $X$ denotes the unknown transformation between the camera and the robot \parencite{tsai1989new, park1994robot}. Various classical methods have been proposed to solve this problem, including the Tsai–Lenz method, Park–Martin method, and Horaud method, which differ in how rotation and translation are estimated and optimized.

\parencite{zeng22icra} configured ROS scan parameters with a translation step size of 10 cm and a rotation step size of 15°, while deep reinforcement learning (DRL) was used to determine the next best action. The environment was segmented into four voxel areas to facilitate efficient exploration and decision-making. However, the approach was tested only in simulation, and significant time was required both for training the DRL model and for performing color thresholding to filter red pixels.

In practical robotic systems, accurate hand-eye calibration is critical for perception-guided manipulation tasks. Errors in calibration can directly affect the alignment between the perceived object position and the robot coordinate system, leading to inaccuracies in motion planning and task execution. These errors may arise from factors such as limited pose diversity during data collection, sensor noise, and imperfect detection of calibration targets.

\subsubsection{Motion Plan}

Motion planning is a fundamental problem in robotics, aiming to compute a feasible and collision-free trajectory that moves a robot from an initial configuration to a goal configuration under various constraints. These constraints may include geometric constraints, kinematic feasibility, dynamic limits, and task-specific requirements. Over the past decades, motion planning approaches have evolved significantly, and can be broadly categorized into graph-based, sampling-based, optimization-based, and learning-based methods \cite{elbanhawi2014sampling, crocker2021review}.

\paragraph{Sampling-based Methods}

Sampling-based methods have become the dominant paradigm for motion planning in high-dimensional spaces. Representative algorithms include the Probabilistic Roadmap (PRM) and Rapidly-exploring Random Tree (RRT). These methods explore the configuration space through random sampling without explicit discretization, enabling efficient planning in complex environments. Variants such as RRT* further improve path quality by introducing asymptotic optimality \cite{karaman2011sampling}. These approaches are widely adopted in practical robotic systems and are implemented in frameworks such as MoveIt via the Open Motion Planning Library (OMPL) \cite{sucan2012ompl}.

\paragraph{Other Approaches}

Graph-based motion planning methods, such as A*, discretize the configuration space and search for an optimal path, but suffer from poor scalability in high-dimensional spaces. Optimization-based approaches, including CHOMP, STOMP, and TrajOpt, refine trajectories through cost minimization but may converge to local minima \cite{ratliff2009chomp, kalakrishnan2011stomp, schulman2014trajopt}. Learning-based approaches, such as reinforcement learning and diffusion policies, offer promising results in complex environments but require substantial training data \cite{chi2023diffusion}.

\paragraph{Challenges and Trends}

Despite significant progress, motion planning remains challenging due to high-dimensional configuration spaces, dynamic environments, and uncertainty in perception. Recent research trends focus on hybrid approaches that combine sampling-based planning with trajectory optimization, as well as perception-aware planning strategies that integrate visual feedback into motion generation \cite{crocker2021review}.

\subsection{VLA}

Recent advances in Vision--Language--Action (VLA) models have introduced a new paradigm for robotic manipulation, where perception, reasoning, and control are unified into a single end-to-end learning framework. Unlike traditional robotics pipelines that rely on explicit perception modules, geometric calibration, and motion planning, VLA approaches directly map visual observations to robot actions through data-driven learning.

One representative work is RT-2, which integrates vision-language models with robotic control, enabling robots to generalize across tasks by leveraging large-scale web data. Similarly, OpenVLA and related frameworks extend large language models (LLMs) to action generation by discretizing robot actions into token sequences and learning them jointly with visual features. These approaches demonstrate strong generalization capabilities and the ability to perform complex manipulation tasks with minimal task-specific engineering.

In parallel, diffusion-based policies have emerged as a powerful alternative for visuomotor learning. Methods such as Diffusion Policy \parencite{chi2023diffusion} model robot actions as a conditional generative process, enabling smooth and robust trajectory generation. These models have shown improved performance in handling multimodal action distributions and uncertainty compared to traditional regression-based policies.

Furthermore, recent VLA frameworks introduce a base policy model (often denoted as $\pi_0$), which represents a pretrained vision--language--action model serving as a general-purpose policy backbone. Such models are typically trained on large-scale, heterogeneous datasets spanning multiple tasks and embodiments, as demonstrated in works like RT-2 and Open X-Embodiment \parencite{openx2023}. 

\section{Problem Statement}

Despite recent advances in spectral sensing and robotic systems, existing approaches for fruit quality assessment still face several critical limitations.

\paragraph{\textbf{Limitation 1: Dependence on Destructive or Controlled Sweetness Measurement Methods}} 
Most sweetness measurement techniques rely on destructive sampling using refractometers, which are labor-intensive and unsuitable for large-scale deployment \parencite{smartref}. Although non-destructive sensing methods such as Near-Infrared (NIR) spectroscopy have shown promising results \parencite{R1, wold2024}, their application is often limited to controlled environments and does not easily scale to real-world agricultural settings.

\paragraph{\textbf{Limitation 2: Challenges in Robust Strawberry Detection and Localization}} 
Reliable detection and localization of strawberries in natural environments remain difficult. Vision-based approaches are significantly affected by occlusions, complex backgrounds, and varying illumination conditions, which are common in greenhouse and field environments. Existing methods often struggle to maintain consistent performance under such conditions \parencite{yu2024tactile, lee2024sonicboom}.

\paragraph{\textbf{Limitation 3: Lack of Integrated Perception and Manipulation Frameworks}} 
Current systems typically treat perception, localization, and quality assessment as separate modules, lacking a unified framework that integrates these components into a coherent pipeline. While robotic systems have been explored for fruit inspection and harvesting \parencite{zeng22icra}, many approaches rely on predefined scanning strategies or simulation-based validation, limiting their applicability in real-world scenarios.

Therefore, there is a clear need for an integrated system that can jointly address detection, localization, and non-destructive sweetness estimation, while being robust to environmental variability and calibration uncertainties. Such a system should tightly couple perception and robotic control, enabling adaptive and scalable fruit quality assessment in real-world agricultural settings.

\section{Project Objectives}
\label{sec:objectives}

Based on the limitations identified in existing approaches, this project focuses on the development of an integrated robotic sensing system for automated strawberry detection and non-destructive sweetness estimation. The objectives are organised to correspond to the key challenges discussed previously.

\subsection{Objective 1: Robust Strawberry Detection and Localization}

\textbf{Related limitation:} Challenges in robust detection and localization under complex environmental conditions.

Strawberry detection in real-world agricultural environments is affected by factors such as occlusion, illumination variation, and background clutter. To address these challenges, this project develops a detection and localization pipeline that combines multiple perception strategies.

Specifically, a deep learning-based detector (YOLO) is employed to provide reliable object-level localisation, while HSV-based segmentation is used to refine object boundaries. In addition, RGB--ToF data fusion is incorporated to improve spatial consistency between visual and depth information. This enables the extraction of stable 2D reference points and their subsequent projection into 3D space.

\subsection{Objective 2: Closed-Loop Robotic Perception and Control}

\textbf{Related limitation:} Lack of integrated robotic manipulation frameworks.

Existing approaches often treat perception and actuation as separate components, limiting system-level performance. In this project, a closed-loop robotic perception and control pipeline is established to integrate sensing, localisation, and motion execution.

An accurate spatial relationship between the camera and the robotic arm is first established through hand--eye calibration. Based on this transformation, detected strawberry positions are mapped into the robot coordinate frame. A waypoint-based search strategy is implemented to support efficient exploration of the workspace, while a closed-loop refinement mechanism allows the robot to iteratively adjust its pose using visual feedback.

Motion execution is achieved through inverse kinematics and trajectory planning, enabling the robotic arm to approach target strawberries in a stable and controlled manner.

\subsection{Objective 3: Non-Destructive Sweetness Estimation}

\textbf{Related limitation:} Dependence on destructive or controlled measurement methods.

Conventional sweetness measurement methods typically require physical sampling or controlled environments. To enable in-situ assessment, this project investigates a non-destructive sweetness estimation approach based on Time-of-Flight (ToF) amplitude signals.

The approach leverages the relationship between optical reflectance characteristics and internal fruit composition. A data-driven model is used to estimate sweetness levels (in $^\circ$Brix) from amplitude features, enabling contact-free prediction.

To improve measurement reliability, the robotic system actively controls the sensor-to-fruit distance, maintaining an optimal range (approximately 15--20 cm) during data acquisition.

\subsection{System Integration Objective}

In addition to the individual components described above, the project aims to integrate detection, localisation, robotic control, and sweetness estimation into a unified system.

The resulting framework forms a closed-loop robotic perception system, in which sensing, decision-making, and actuation are tightly coupled. This integration enables automated and scalable fruit quality assessment in practical agricultural environments.

\section{Methodology and Implementation}
\subsection{System Architecture}
\label{sec:system_architecture}

\subsubsection{Overview}
The proposed strawberry manipulation system is organised as a layered architecture comprising
four functional stages: sensing, perception and metric localisation, motion planning, and
low-level joint execution.

\begin{figure}[H]
  \centering
  \includegraphics[width=\linewidth]{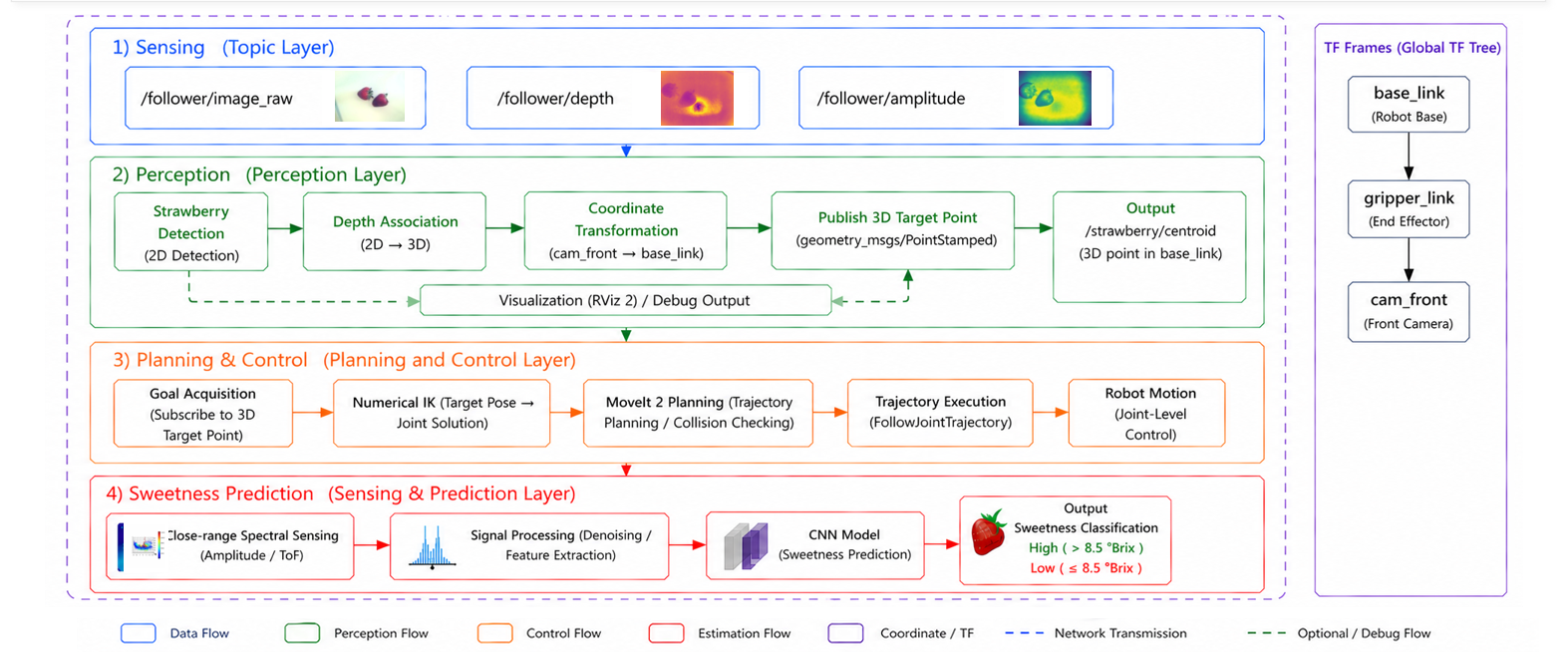}
  \caption{Overview of the robotic system for strawberry localisation and non-destructive sweetness prediction, including sensing, perception, planning and control, and learning-based estimation modules.}
  \label{fig:system_architecture}
\end{figure}

\begin{figure}[H]
  \centering
  \fbox{%
    \parbox{0.92\linewidth}{%
      \centering
      \includegraphics[width=\linewidth]{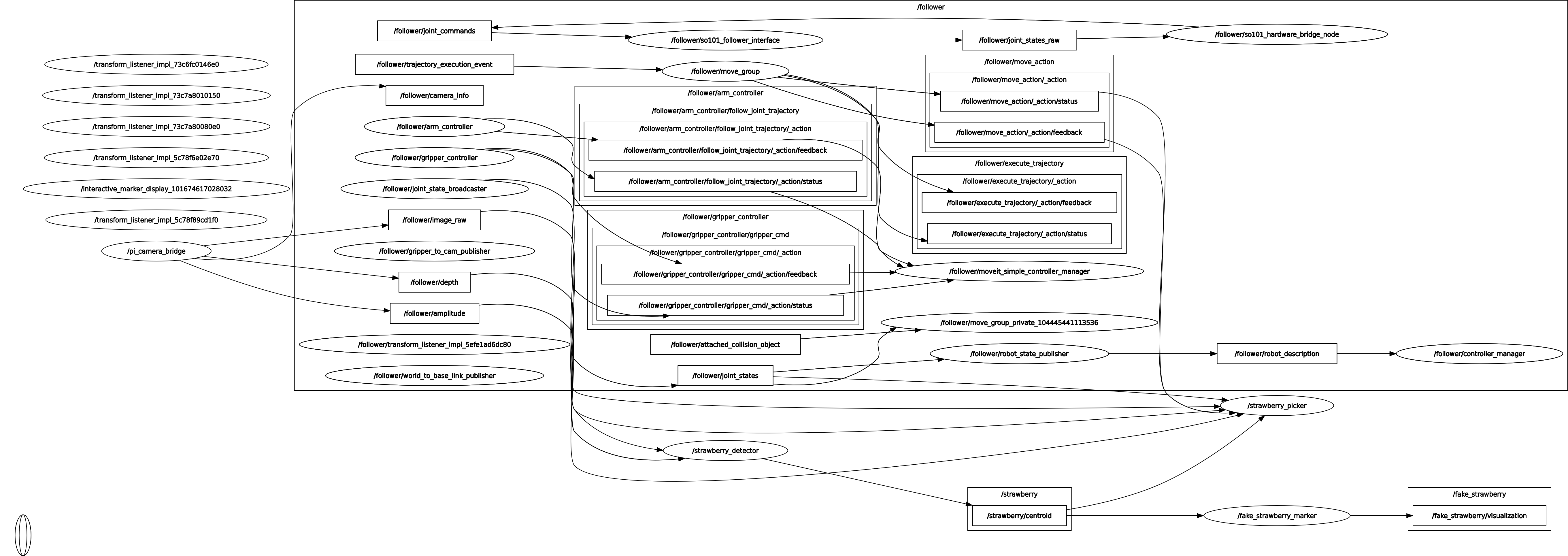}
    }%
  }
  \caption{ROS~2 computation graph of the integrated strawberry manipulation pipeline,
  showing node-level interactions, topic communication, and action interfaces.}
  \label{fig:ros_computation_graph}
\end{figure}

Physically, an RGB--ToF camera module is mounted on the wrist of an SO101-class follower arm
(eye-in-hand configuration), while joint commands are issued through a ROS~2 controller interface. Figure~\ref{fig:system_architecture} provides a high-level view of the data flow and module
interactions, whereas Figure~\ref{fig:ros_computation_graph} details the underlying ROS~2
computation graph, including nodes, topics, and action interfaces.
The following sections describe each layer in accordance with the runtime data flow.

\subsubsection{Sensing layer}
The sensing layer is responsible for streaming multi-modal visual data from the wrist-mounted
camera system.
A bridge node publishes synchronised RGB, ToF depth, and
amplitude images topics, including
\texttt{/follower/image\_raw}, \texttt{/follower/depth}, \\
and \texttt{/follower/amplitude}.
All image streams are expressed in the camera coordinate frame (\texttt{cam\_front}) and are
time-stamped for downstream synchronization.

In parallel, the robot driver stack publishes joint states via
\texttt{/follower/joint\_states} and maintains controller-level interfaces
\texttt{/follower/joint\_commands}.
These signals ensure consistency between the physical robot configuration, the TF tree, and
the planning scene used by MoveIt~2.

\subsubsection{Perception and localisation layer}
The perception layer transforms raw visual observations into a geometrically meaningful
manipulation target.
A dedicated node (\ \texttt{/strawberry\_detector}) subscribes to the RGB and depth streams to perform strawberry detection and segmentation.

The detected 2D region is fused with ToF depth measurements to estimate the 3D centroid of
the target.
The resulting position is published as a \texttt{geometry\_msgs/PointStamped} message on
\texttt{/strawberry/centroid}, expressed in the \texttt{base\_link} frame via TF-based
coordinate transformation.

This layer does not perform motion generation; instead, it provides a continuously updated
Cartesian goal for downstream planning.
Optional debug outputs, including annotated images and RViz markers, are published to support
runtime inspection and system validation.

\subsubsection{Planning and execution layer}
The planning layer consumes the target position and generates executable robot motions.
A high-level control node (\ \texttt{/strawberry\_picker}) subscribes to
\texttt{/strawberry/centroid} and computes a feasible end-effector pose.

Inverse kinematics is solved numerically to obtain joint-space solutions, which are then
passed to MoveIt~2 (\texttt{/move\_group}) for collision-aware trajectory planning.
Planned trajectories are executed via standard ROS~2 control interfaces, including
\texttt{FollowJointTrajectory} actions provided by the arm and gripper controllers.

The execution stack includes components such as
\texttt{/arm\_controller}, \\
\texttt{/gripper\_controller}, and
\texttt{/joint\_state\_broadcaster}, which interface directly with the robot hardware.

To improve robustness against perception noise and calibration error, the system adopts an
incremental closed-loop strategy: short motion segments are executed, followed by re-perception,
thereby reducing error accumulation over long trajectories.

\subsubsection{Transform tree}
A consistent transform (TF) tree is maintained throughout the system, rooted at
\texttt{base\_link}.
Unlike conventional fixed camera setups, the eye-in-hand configuration introduces a dynamic
relationship between the robot and the sensor.

The TF hierarchy follows:
\begin{center}
\texttt{base\_link} $\rightarrow$ \texttt{gripper\_link} $\rightarrow$ \texttt{cam\_front}
\end{center}

Two key extrinsic calibration problems are solved offline and used as static transforms:
\begin{itemize}
    \item \textbf{RGB--ToF calibration:} Intrinsic parameters and stereo extrinsics obtained
    via calibration (circle-grid), enabling consistent depth fusion.
    
    \item \textbf{Hand--eye calibration:} Rigid transform between
    \texttt{gripper\_link} and \texttt{cam\_front}, defining the camera pose relative to the
    end-effector.
\end{itemize}

These static transforms are combined with the time-varying transform from
\texttt{base\_link} to \texttt{gripper\_link}, obtained from joint encoders and published via
the robot state publisher.
Together, they form a coherent spatial representation that enables accurate mapping from
image coordinates to the robot workspace for motion planning.
\subsection{Hardware Setup}
\label{sec:hardware_setup}

The robotic system integrates an RGB camera and a ToF sensor mounted on a SO-101 robotic arm,
enabling flexible positioning and multi-angle data acquisition for strawberry observation.
This section describes both the physical mounting design and the hand--eye calibration
procedure required to align perception with the robot coordinate system.

\subsubsection{3D-Printed Sensor Mount}
\paragraph{Design and Implementation}

\begin{figure}[H]
    \centering
    \begin{subfigure}[b]{0.32\textwidth}
        \centering
        \includegraphics[width=\textwidth]{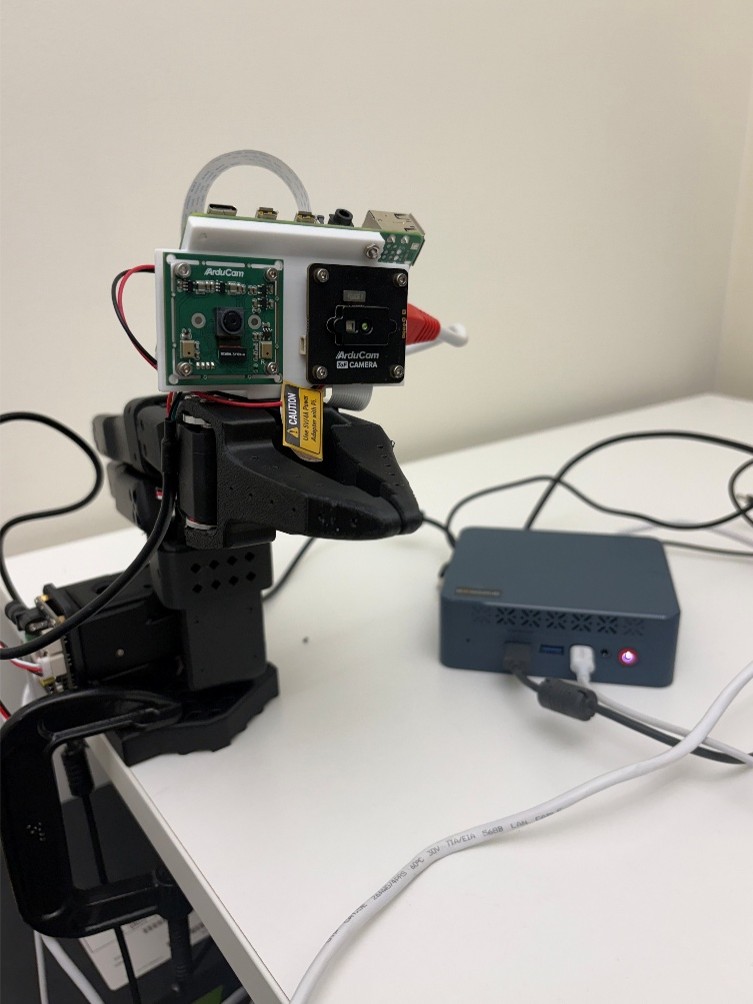}
        \caption{Front view}
        \label{fig:front}
    \end{subfigure}
    \hfill
    \begin{subfigure}[b]{0.32\textwidth}
        \centering
        \includegraphics[width=\textwidth]{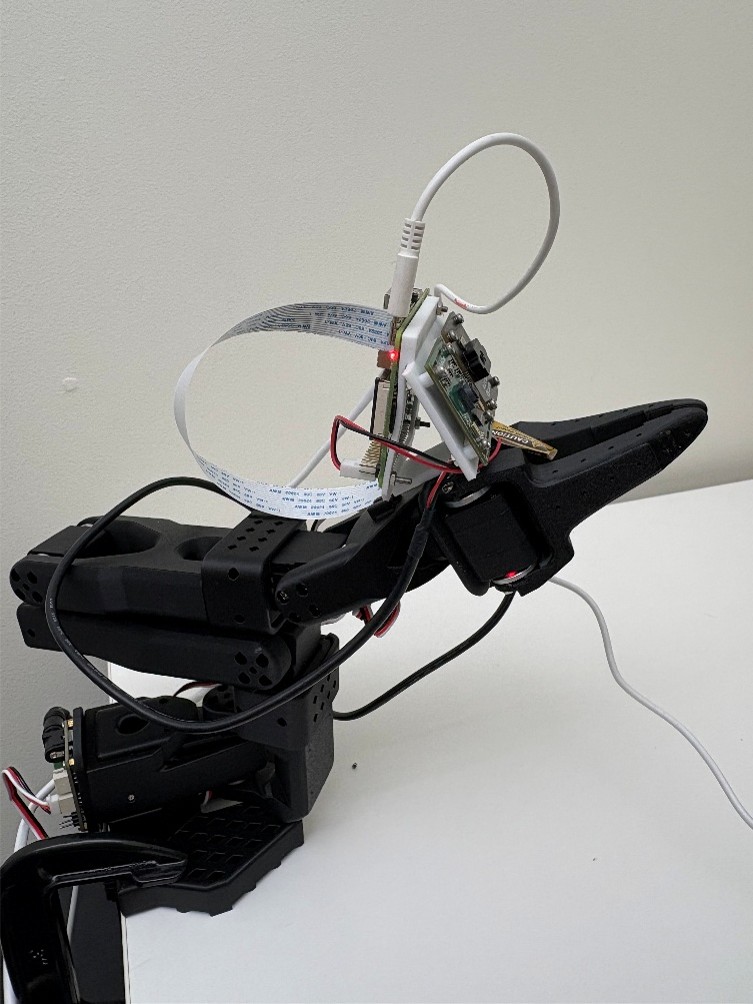}
        \caption{Side view}
        \label{fig:side}
    \end{subfigure}
    \hfill
    \begin{subfigure}[b]{0.32\textwidth}
        \centering
        \includegraphics[width=\textwidth]{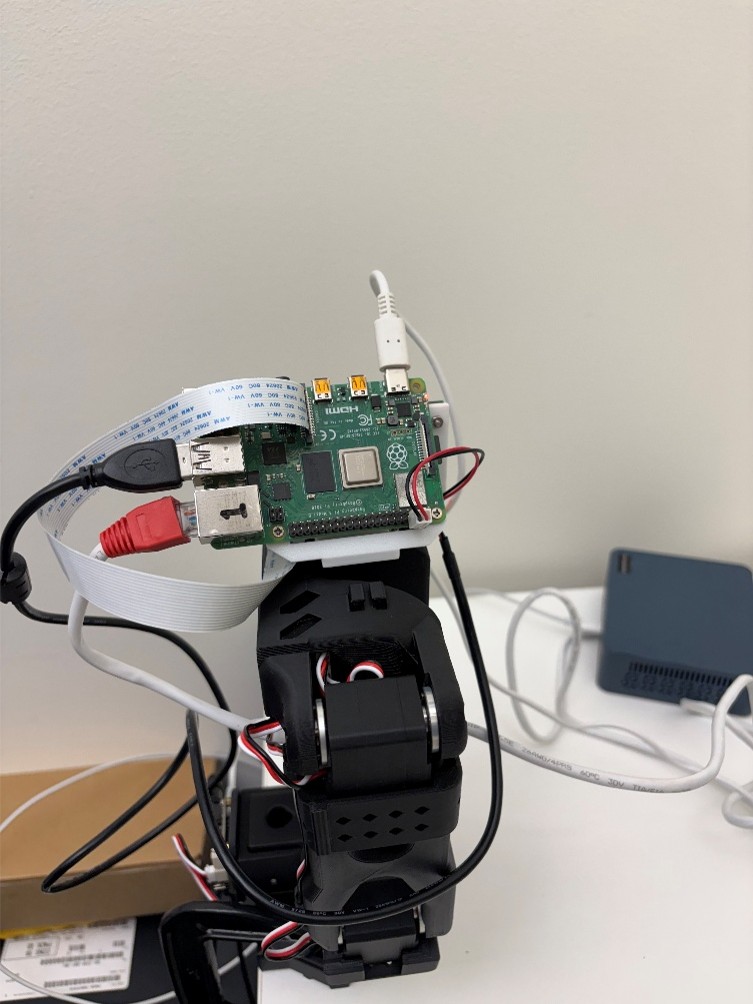}
        \caption{Back view}
        \label{fig:back}
    \end{subfigure}

    \caption{3D-printed sensor mount (front, side, and back views)}
    \label{fig:mount_views}
\end{figure}

To facilitate sensor integration, a custom 3D-printed mount shown in
Figure~\ref{fig:mount_views} was designed and fabricated for this project.

The key design features are summarised as follows:

\begin{itemize}
    \item \textbf{Lightweight and stable:} The mount is lightweight while maintaining high mechanical stability, minimizing additional load on the robotic arm and reducing camera vibration during motion.
    
    \item \textbf{Low-cost fabrication:} The structure is fully self-designed and produced via 3D printing (approximately AUD 5 per component), enabling rapid prototyping and iteration.
    
    \item \textbf{Adaptable design:} The mount supports flexible modification for different sensor configurations and robotic platforms, ensuring versatility for future experiments and deployments.
    
    \item \textbf{Sensor alignment:} A 0.5\,cm elevation is introduced to compensate for electronic components behind the sensors, ensuring both RGB and ToF sensors lie approximately on the same plane.
    
    \item \textbf{Rigid attachment:} The sensors are firmly fixed to the end-effector, enhancing structural rigidity and ensuring stable positioning during operation.
\end{itemize}

\subsubsection{Hand--Eye Calibration}
\label{sec:handeye_calibration}

The camera is mounted on the wrist (eye-in-hand configuration), requiring estimation of a
fixed rigid transformation between the robot end-effector frame \texttt{gripper\_link} and
the camera frame \texttt{cam\_front}.
This transformation enables detected 3D points in the camera frame to be accurately
transformed into the robot base frame \texttt{base\_link} for motion planning.

\paragraph{Calibration Strategy}

Two approaches were considered.
\paragraph{MoveIt~2 calibration (not adopted)}
The MoveIt~2 Calibration workflow~\parencite{orsula2023moveit2calibration} provides a
convenient GUI-based pipeline, but it was not well suited for this setup due to limited
control over pose selection and reduced transparency in the solving process.
These limitations made debugging and iteration difficult for a custom RGB--ToF system.
In addition, when directly using joint states from the simulation environment, noticeable
discrepancies were observed compared to the real robot configuration, leading to relatively
large calibration errors. This further reduced the reliability of the estimated hand--eye
transformation in our experimental setup.

\paragraph{Custom hand--eye calibration (adopted)}
Instead, a manual data collection approach was used.
Robot forward kinematics provides the gripper pose relative to \texttt{base\_link}, while
a ChArUco board observed in the image provides the target pose relative to
\texttt{cam\_front}.
These paired samples are used to estimate the hand--eye transformation offline, which is
then published to the ROS~2 \texttt{/tf} tree.

This approach is preferred because it is:

\begin{itemize}
    \item \textbf{Transparent:} Individual samples can be inspected and filtered
    \item \textbf{Flexible:} Pose selection and thresholds can be tailored to the hardware
    \item \textbf{Lightweight:} Easily integrated into the existing ROS~2 pipeline
\end{itemize}

ChArUco boards are preferred over single ArUco markers due to their higher corner density
and improved robustness under partial occlusion.

\begin{figure}[htbp]
    \centering
    \begin{subfigure}{0.45\textwidth}
        \centering
        \includegraphics[width=\linewidth]{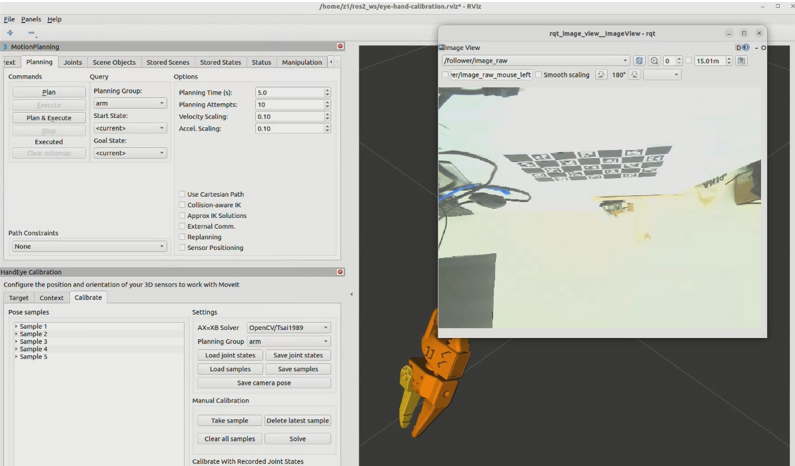}
        \caption{MoveIt~2 calibration GUI}
        \label{fig:moveit_handeye_gui}
    \end{subfigure}
    \hfill
    \begin{subfigure}{0.45\textwidth}
        \centering
        \includegraphics[width=\linewidth]{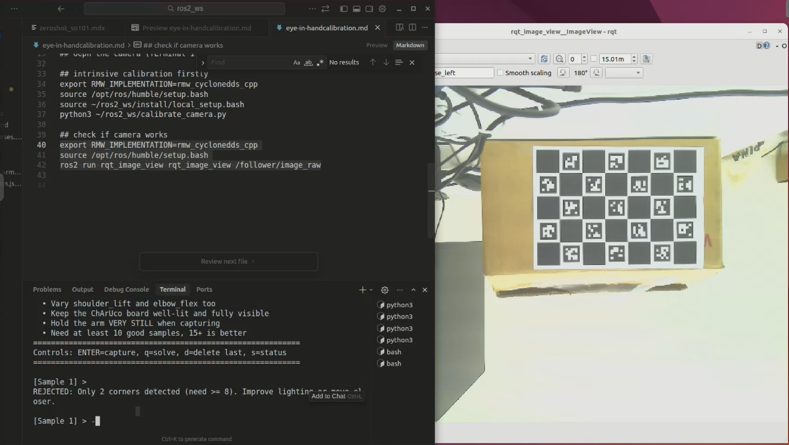}
        \caption{Custom OpenCV workflow}
        \label{fig:custom_handeye_workflow}
    \end{subfigure}
    \caption{Comparison of hand--eye calibration approaches.}
    \label{fig:handeye_two_approaches}
\end{figure}

\begin{figure}[H]
\centering
\begin{subfigure}[b]{0.45\textwidth}
    \centering
    \includegraphics[width=\textwidth]{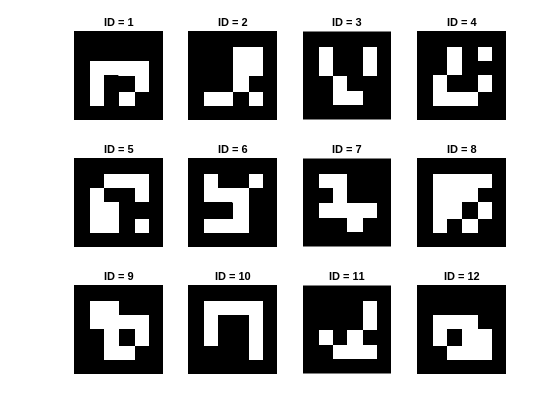}
    \caption{ArUco board}
    \label{fig:aruco}
\end{subfigure}
\hfill
\begin{subfigure}[b]{0.45\textwidth}
    \centering
    \includegraphics[width=\textwidth]{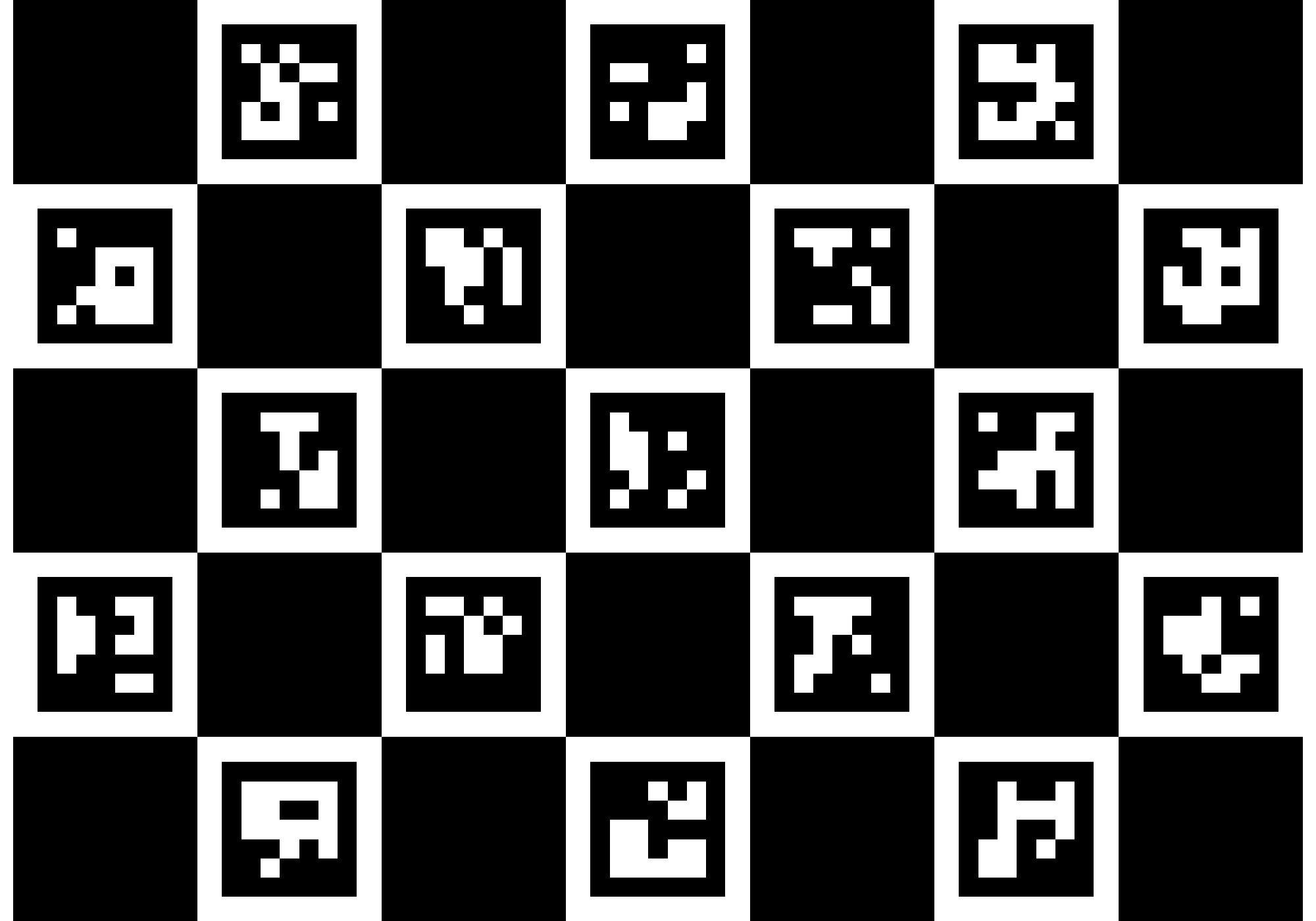}
    \caption{ChArUco board}
    \label{fig:charuco}
\end{subfigure}
\caption{Comparison of calibration targets used in this project}
\label{fig:aruco_charuco}
\end{figure}

\paragraph{Calibration Target Selection (ArUco vs ChArUco)}
Accurate camera and hand--eye calibration require reliable and geometrically informative feature correspondences. In this work, two commonly used calibration targets, namely ArUco boards and ChArUco boards, are considered and compared. An ArUco board consists of multiple independent square markers encoded with unique binary patterns, enabling robust detection and identification even under partial occlusion. However, each marker provides only four corner points, resulting in a relatively sparse set of feature correspondences and limiting calibration accuracy. In contrast, a ChArUco board combines ArUco markers with a chessboard structure, allowing additional corner points to be interpolated across the grid. This hybrid design preserves the robustness of marker-based detection while significantly increasing the number of available feature points, thereby improving geometric consistency and pose estimation accuracy. A comparison between the two calibration targets is summarised in Table~\ref{tab:aruco_charuco_comparison}, and representative examples are shown in Fig.~\ref{fig:aruco_charuco}. As can be observed, the ChArUco board provides a denser and more stable set of correspondences, leading to improved calibration precision and reduced sensitivity to noise. Experimental observations in this project further indicate that ChArUco-based calibration achieves lower translation and rotation errors compared to ArUco-based methods. Therefore, considering the requirement for accurate spatial transformation between the camera and the robotic arm, the ChArUco board is selected as the primary calibration target.
\begin{table}[H]
\centering
\caption{Comparison between ArUco and ChArUco calibration boards}
\label{tab:aruco_charuco_comparison}
\begin{tabular}{lcc}
\hline
\textbf{Criterion} & \textbf{ArUco Board} & \textbf{ChArUco Board} \\
\hline
Feature type & Marker corners only & Marker + chessboard corners \\
Number of features & Sparse (4 per marker) & Dense (interpolated grid) \\
Detection robustness & High (ID-based) & High (ID + grid constraint) \\
Calibration accuracy & Moderate & High \\
Stability under occlusion & Strong & Strong \\
Sensitivity to noise & Moderate & Lower \\
Pose estimation quality & Limited & More precise \\
Suitability for hand--eye calibration & Acceptable & Preferred \\
\hline
\end{tabular}
\end{table}

\subsection{Strawberry Detection and Segmentation }
\subsubsection{Circle-Grid Two-Camera Calibration (RGB and ToF)}
\label{sec:rgb_tof_calib}
\paragraph{Why Circle Grid}

In this work, a circle grid calibration pattern is adopted instead of the traditional chessboard pattern for camera calibration. While chessboard-based methods are widely used due to their high corner localization accuracy, they are known to be sensitive to real-world imaging conditions such as motion blur, lighting variations, and partial occlusions. In robotic manipulation scenarios, especially with a moving end-effector and varying viewpoints, these factors can significantly degrade the robustness of corner detection.

In contrast, circle grid patterns rely on blob detection to estimate the centers of circular features, which tends to be more robust under challenging conditions. The absence of sharp edges makes circle detection less sensitive to image blur and illumination changes. Furthermore, asymmetric circle grid configurations provide inherent geometric constraints that improve pattern disambiguation and detection stability across different orientations.

Although the center-based localization in circle grids may offer slightly lower precision compared to subpixel corner detection in chessboards, the improved robustness and detection reliability make circle grids more suitable for real-world robotic calibration tasks. This trade-off is particularly important in our system, where stable calibration under dynamic sensing conditions is prioritised over marginal gains in static accuracy.

\begin{table}[H]
\centering
\caption{Comparison between Chessboard and Circle Grid Calibration Patterns}
\label{tab:circle_vs_chessboard}
\begin{tabular}{lcc}
\hline
\textbf{Aspect} & \textbf{Chessboard} & \textbf{Circle Grid} \\
\hline
Feature type & Corner points & Circular blobs (centers) \\
Localization accuracy & High (subpixel corners) & Moderate (center estimation) \\
Robustness to blur & Low & High \\
Sensitivity to lighting & High & Lower \\
Occlusion tolerance & Low & Moderate \\
Detection stability & Sensitive to viewpoint & More stable \\
Algorithm complexity & Low & Moderate (blob detection) \\
Suitability for robotics & Moderate & High \\
\hline
\end{tabular}
\end{table}
The calibration dataset was collected by recording, for each pose, an RGB image together with the
synchronously triggered ToF amplitude image (and depth where available).
Temporal alignment ensures that each RGB frame has a matching ToF capture for offline processing.

\begin{figure}[H]
    \centering
    \begin{subfigure}[b]{0.32\textwidth}
        \centering
        \includegraphics[width=\textwidth]{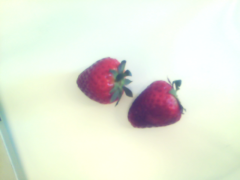}
        \caption{RGB}
        \label{fig:strawberry_rgb}
    \end{subfigure}
    \hfill
    \begin{subfigure}[b]{0.32\textwidth}
        \centering
        \includegraphics[width=\textwidth]{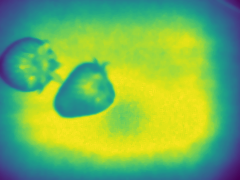}
        \caption{ToF amplitude}
        \label{fig:strawberry_amplitude}
    \end{subfigure}
    \begin{subfigure}[b]{0.32\textwidth}
        \centering
        \includegraphics[width=\textwidth]{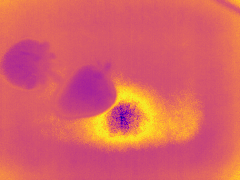}
        \caption{ToF depth}
        \label{fig:strawberry_depth}
    \end{subfigure}
    \caption{Example RGB, ToF amplitude, and ToF depth streams used in this work.}
    \label{fig:strawberry_overview}
\end{figure}

\begin{figure}[H]
    \centering
    \begin{subfigure}[b]{0.32\textwidth}
        \centering
        \includegraphics[width=\textwidth]{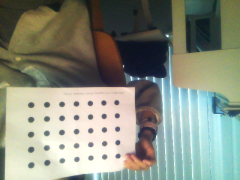}
        \caption{RGB}
        \label{fig:rgb_circlegrid}
    \end{subfigure}
    \hfill
    \begin{subfigure}[b]{0.32\textwidth}
        \centering
        \includegraphics[width=\textwidth]{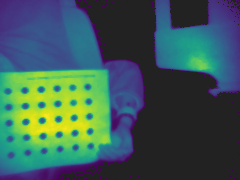}
        \caption{ToF amplitude}
        \label{fig:tof_amplitude_circlegrid}
    \end{subfigure}
    \hfill
    \begin{subfigure}[b]{0.32\textwidth}
        \centering
        \includegraphics[width=\textwidth]{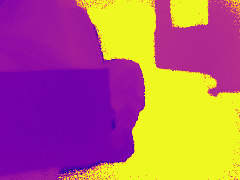}
        \caption{ToF depth}
        \label{fig:tof_depth_circlegrid}
    \end{subfigure}
    \caption{Example circle-grid images captured by the RGB and ToF sensors.}
    \label{fig:circlegrid_comparison}
\end{figure}

Geometric calibration between the RGB camera and the ToF module used the \emph{amplitude} images
to detect a printed \emph{circle grid}: circle centres appear with much higher contrast than the
same pattern in raw depth, where the board is often hard to see for a standard flat print.
Figure~\ref{fig:circlegrid_comparison} compares the three modalities for a typical calibration pose.

\begin{figure}[H]
    \centering
    \resizebox{0.8\textwidth}{!}{%
    \begin{subfigure}[b]{0.45\textwidth}
        \centering
        \includegraphics[width=\textwidth]{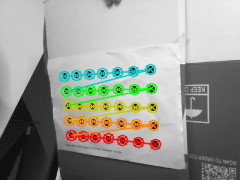}
        \caption{RGB circle-grid detection}
        \label{fig:calib_rgb_085}
    \end{subfigure}
    \hfill
    \begin{subfigure}[b]{0.45\textwidth}
        \centering
        \includegraphics[width=\textwidth]{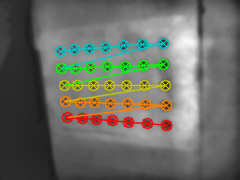}
        \caption{ToF amplitude circle-grid detection}
        \label{fig:calib_amp_085}
    \end{subfigure}
    } 
    \caption{Representative successful detections used in calibration.}
    \label{fig:combined_calibration}
\end{figure}

A total of 30 poses were recorded with the circle grid at varied distances and orientations
(Figure~\ref{fig:combined_calibration}).
Circle centres were detected in \emph{both} modalities for 20 of the 30 captures using
\texttt{cv2.findCirclesGrid}.
Those 20/30 pairs were used to estimate intrinsics for each view (focal length, principal point,
and distortion) and the extrinsic transform (rotation and translation) from the ToF frame to
the RGB frame, with \texttt{cv2.stereoCalibrate} while fixing the previously estimated intrinsics.

Table~\ref{tab:calibration_set2} summarises the calibration quality.
The RGB RMS reprojection error is 0.156~px; the ToF (amplitude) RMS is 0.284~px, slightly larger
as expected given lower resolution and stronger noise, but still indicating a stable fit between
observed circle centres and the calibrated projection model.

\begin{table}[H]
\centering
\caption{Calibration results (object-space units: millimetres).}
\resizebox{0.8\textwidth}{!}{%
\begin{tabular}{|l|c|}
\hline
\textbf{Item} & \textbf{Value} \\
\hline
RGB RMS reprojection error & 0.156~px \\
\hline
ToF (amplitude) RMS reprojection error & 0.284~px \\
\hline
Rotation (ToF $\rightarrow$ RGB) &
$\begin{bmatrix} 0.9981 & -0.0162 & -0.0592 \\ 0.0190 & 0.9987 & 0.0471 \\ 0.0584 & -0.0481 & 0.9971 \end{bmatrix}$ \\
\hline
Translation (ToF $\rightarrow$ RGB) & $[-2.707,\,-0.243,\,0.947]$ \\
\hline
Valid image pairs used & 20/30 \\
\hline
\end{tabular}
} 
\label{tab:calibration_set2}
\end{table}

The rotation matrix is close to the identity, indicating that the RGB and ToF coordinate axes
are nearly parallel with only a small relative tilt.
The translation vector is expressed in the same millimetre units as the circle spacing used in
\texttt{square\_size}; its components are on the order of a few millimetres (not sub-millimetre),
reflecting the physical baseline and mounting offset between the two sensors.

The remaining 10/30 captures were discarded when circle detection failed in at least one modality
due to occlusion, motion blur, specularities, or insufficient contrast.

Overall, these results provide a consistent RGB--ToF geometric model and a reasonable number of
independent constraints (20/30), forming the basis for subsequent depth linking and 3D fusion.

\begin{figure}[H]
    \centering
    \resizebox{0.8\textwidth}{!}{%
    \begin{subfigure}[b]{0.45\textwidth}
        \centering
        \includegraphics[width=\textwidth]{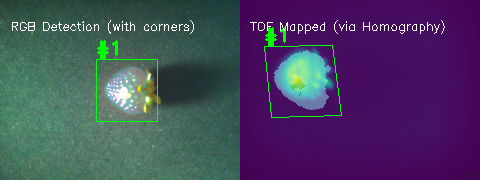}
        \caption{Single strawberry}
        \label{fig:val_strawberry_one}
    \end{subfigure}
    \hfill
    \begin{subfigure}[b]{0.45\textwidth}
        \centering
        \includegraphics[width=\textwidth]{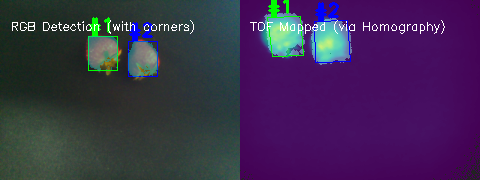}
        \caption{Two strawberries}
        \label{fig:val_strawberry_two}
    \end{subfigure}
    } 
    \caption{Qualitative validation on strawberry scenes after calibration.}
    \label{fig:calibration_validation_strawberries}
\end{figure}

Calibration quality was further checked by qualitatively overlaying aligned RGB and depth-related
views on strawberry scenes (Figure~\ref{fig:calibration_validation_strawberries}).

\subsubsection{Detection Model}
As shown in Table~\ref{tab:yolo_retinanet}, YOLOv11s significantly outperforms RetinaNet in inference speed, achieving approximately 20$\times$ faster processing on CPU. While RetinaNet provides higher detection precision, its latency makes it unsuitable for real-time robotic applications.

\begin{table}[h]
\centering
\caption{RetinaNet vs YOLOv11s for Real-Time Strawberry Detection}
\small
\begin{tabular}{lcc}
\hline
\textbf{Metric} & \textbf{RetinaNet} & \textbf{YOLOv11s} \\
\hline
Detection Accuracy & High & Moderate--High \\
Small Dataset Performance & Strong (Focal Loss) & Good (Fine-tuned) \\
Inference Speed (CPU) & $\sim$5000 ms & $\sim$250 ms \\
Frame Rate & $\sim$0.2 FPS & $\sim$4 FPS \\
Real-time Capability & $\times$ & $\checkmark$ \\
False Positives & Low & Slightly Higher \\
Robustness (Hand vs Berry) & High & High \\
Deployment Suitability & Offline & Real-time Robotics \\
\hline
\textbf{Primary Advantage} & Accuracy & Speed \\
\hline
\end{tabular}
\label{tab:yolo_retinanet}
\end{table}

This comparison reflects the fundamental trade-off between detection accuracy and inference latency in real-time perception systems. RetinaNet, with its focal loss design, is better suited for accuracy-critical tasks and performs well on small datasets. However, its computational cost limits its applicability in time-sensitive scenarios.

In contrast, YOLOv11s achieves a more balanced performance by significantly reducing inference time while maintaining acceptable detection accuracy. This makes it more suitable for closed-loop robotic systems, where perception and control must operate in near real-time.  

Therefore, a multi-strategy detection framework was designed to ensure robustness under varying environmental conditions and real-time constraints. The system supports three segmentation modes: HSV-based colour segmentation, YOLOv11s-based object detection, and ToF amplitude-based region growing.

Initially, RetinaNet was considered due to its strong performance on small datasets, particularly with its focal loss formulation that mitigates class imbalance. However, its inference speed (~5000 ms per frame on CPU) makes it unsuitable for real-time robotic applications.

To address this limitation, a YOLOv11s model fine-tuned on the strawberry dataset was adopted. The model achieves significantly faster inference (~250 ms per frame on CPU), enabling near real-time operation. However, experiments revealed that its performance degrades under low-resolution input (240×180), occasionally leading to false positives (e.g., detecting the operator’s hand as a strawberry) and missed detections for small objects.

To improve robustness, an HSV-based segmentation pipeline was implemented as a complementary approach. This method leverages carefully tuned colour thresholds combined with shape-based filtering (e.g., solidity, aspect ratio, and extent) to effectively distinguish strawberries from background and human hands in controlled environments.

Additionally, a ToF amplitude-based segmentation method was incorporated to exploit depth confidence information, enabling detection independent of RGB appearance.

These approaches together form a hybrid detection framework, where YOLOv11s provides semantic detection capability, HSV segmentation offers a fast and reliable fallback under controlled conditions, and ToF-based segmentation enhances robustness in challenging scenarios.

This design reflects a practical trade-off between accuracy, robustness, and real-time performance in robotic perception systems.

\subsubsection{Segmentation Model}

A custom RGB–ToF segmentation approach is used instead of relying on existing off-the-shelf segmentation models. Although many modern segmentation frameworks (e.g., Mask-based CNNs and transformer-based models) can generate object masks, these masks are typically produced at a lower-resolution feature level and then upsampled, which limits their per-pixel accuracy. For this application, precise segmentation is crucial because the ToF camera is a specialized depth sensor that not only provides dense depth information but also outputs amplitude and phase measurements that describe the strength and characteristics of the returned infrared signal. These additional ToF signals can later be exploited to predict strawberry sweetness, as amplitude and phase behavior correlate with surface reflectance and subsurface scattering. To ensure  these ToF measurements can be reliably linked to the exact strawberry surface, the segmentation mask must be perfectly aligned with the calibrated ToF pixels—an accuracy that generic segmentation models cannot consistently provide, especially given the limited dataset and varying outdoor lighting. Therefore, an RGB–ToF calibration and segmentation pipeline offers higher geometric fidelity and better supports downstream tasks such as sweetness estimation.

The segmentation pipeline consists of four key stages designed to ensure accurate alignment between the RGB-derived strawberry mask and the ToF amplitude data. 

\begin{figure}[H]
\centering

\begin{subfigure}[b]{0.32\textwidth}
    \includegraphics[width=\linewidth]{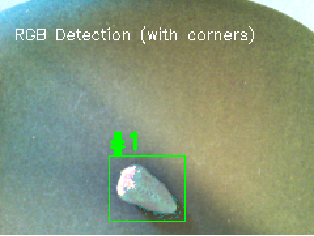}
    \caption{RGB original image}
    \label{fig:rgb_original}
\end{subfigure}
\begin{subfigure}[b]{0.32\textwidth}
    \includegraphics[width=\linewidth]{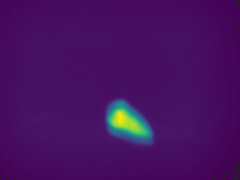}
    \caption{ToF amplitude image}
    \label{fig:tof_amplitude}
\end{subfigure}
\begin{subfigure}[b]{0.32\textwidth}
    \includegraphics[width=\linewidth]{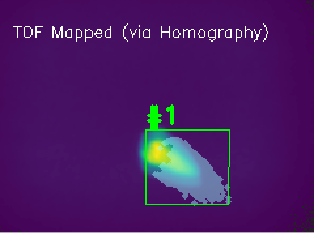}
    \caption{RGB projected to ToF}
    \label{fig:rgb_projected}
\end{subfigure}

\vspace{0.3cm}

\begin{subfigure}[b]{0.32\textwidth}
    \includegraphics[width=\linewidth]{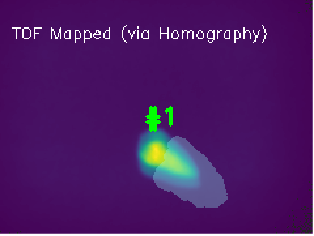}
    \caption{Denoising and CLAHE}
    \label{fig:denoise_clahe}
\end{subfigure}
\begin{subfigure}[b]{0.32\textwidth}
    \includegraphics[width=\linewidth]{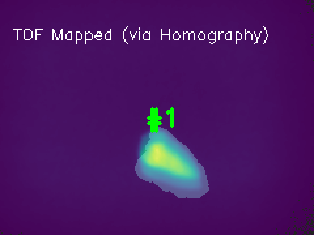}
    \caption{Parameter tuning}
    \label{fig:parameter_tuning}
\end{subfigure}
\begin{subfigure}[b]{0.32\textwidth}
    \includegraphics[width=\linewidth]{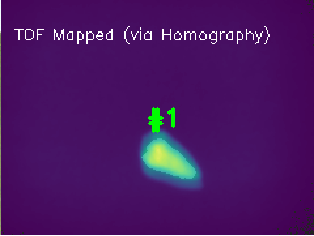}
    \caption{Intersection with ToF}
    \label{fig:intersection}
\end{subfigure}

\caption{Visualization of the strawberry segmentation pipeline.}
\label{fig:segmentation_pipeline}
\end{figure}

\paragraph{Projection via rotation and translation matrix \ref{fig:rgb_projected}}
The binary segmentation mask from the RGB image is geometrically projected into the ToF coordinate frame so that each pixel in the ToF image corresponds to the same physical point on the strawberry surface. This step ensures spatial consistency between the two sensing modalities.

\paragraph{Denoising and CLAHE \ref{fig:denoise_clahe}}
The raw ToF signal often contains speckle noise and low-contrast regions due to sensor physics and varying reflectance. A denoising filter suppresses high-frequency noise, while Contrast-Limited Adaptive Histogram Equalization (CLAHE) enhances local contrast. Together, these operations produce a smoother and more consistent strawberry contour in the amplitude domain, which improves the reliability of later segmentation steps.

\paragraph{Parameter Tuning \ref{fig:parameter_tuning}}
 While segmentation thresholds and filter settings may also be adjusted, the main purpose of this stage is to correct spatial offsets introduced during calibration, ensuring that the projected RGB mask accurately overlaps with the corresponding strawberry region in the ToF image. This adjustment improves the positional consistency of the mask before the two modalities are combined.

\paragraph{Intersection with ToF mask \ref{fig:intersection}}
After preprocessing both modalities, the projected RGB segmentation mask and the enhanced ToF amplitude response are combined using a logical intersection. This step focuses the mask onto the high-intensity ToF regions corresponding to the strawberry surface, removing mismatched pixels and localizing the segmentation more precisely. The resulting mask provides pixel-accurate coverage and is well suited for extracting ToF amplitude profiles for subsequent sweetness estimation.

\subsubsection{Strawberry Localization}
\label{sec:localization_base}
\begin{figure}[H]
\centering
\includegraphics[width=\textwidth]{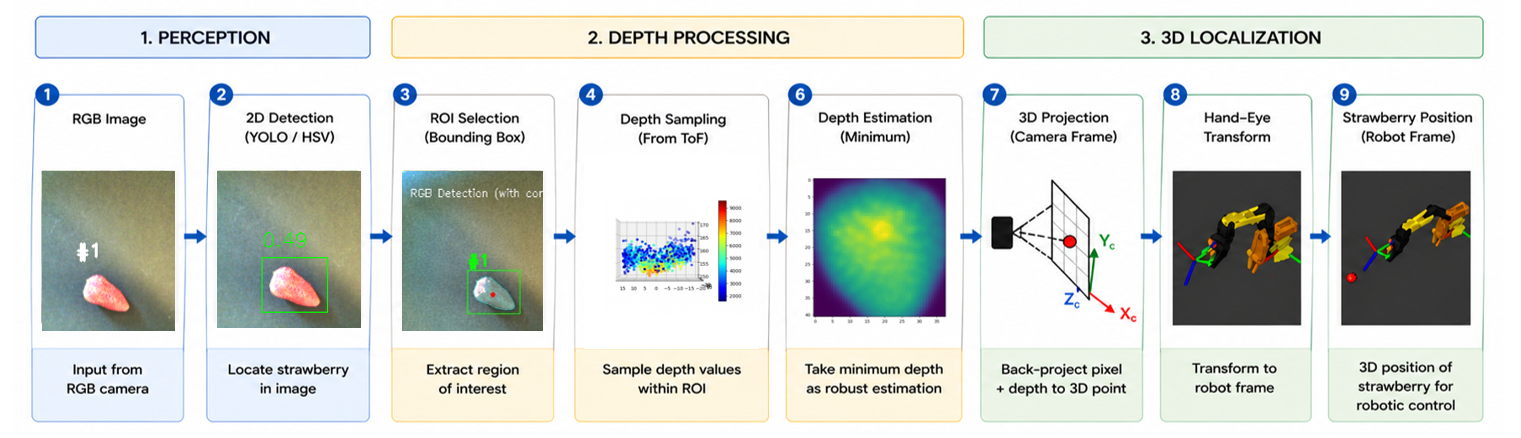}
\caption{Pipeline for strawberry perception, depth processing, and 3D localisation. The system integrates RGB-based detection, ToF depth sampling, and coordinate transformation to estimate the strawberry position in the robot frame.}
\label{fig:3d_localization_pipeline}
\end{figure}

\paragraph{3D Localisation Pipeline}
Figure~\ref{fig:3d_localization_pipeline} illustrates the overall pipeline for strawberry localisation. The process begins with RGB-based perception, where a 2D detector (YOLO or HSV-based segmentation) is used to identify the strawberry in the image and define a region of interest (ROI). Depth information from the Time-of-Flight (ToF) sensor is then sampled within the ROI, and a robust depth estimate is obtained using a minimum-depth or nearest-surface strategy. 

The 2D image coordinates are subsequently back-projected into 3D space to obtain the strawberry position in the camera frame. Finally, a hand--eye calibration transformation is applied to map the position into the robot coordinate frame, enabling downstream robotic control and motion planning. This pipeline forms the foundation of the closed-loop perception and manipulation system.
\paragraph{2D Detection}
Two-dimensional localization is performed in a cascaded manner to combine the semantic
robustness of a detector with the geometric precision of colour-based segmentation.
The overall pipeline consists of the following steps:

\begin{enumerate}
    \item \textbf{Object Detection (ROI Extraction).} 
    A YOLO-style object detector is applied to the full-resolution RGB frame to obtain
    one or more axis-aligned candidate boxes 
    $\mathcal{B}=\{[x_1,y_1,x_2,y_2]\}$.
    Each box defines a region of interest (ROI), which suppresses irrelevant background
    structures and constrains subsequent processing to a compact image area.

    \item \textbf{Colour Segmentation (HSV Thresholding).} 
    For each ROI, the corresponding image patch is extracted (optionally expanded by a
    small margin to avoid cutting the fruit boundary) and converted to the HSV colour space.
    Red pixels are selected using two hue intervals to account for hue wrap-around near
    $H{=}0$, producing a binary mask that is refined with morphological closing and opening
    to reduce noise.

    \item \textbf{Contour Extraction and Filtering.} 
    Outer contours are extracted from the binary mask. Spurious components are removed
    using standard size and shape heuristics (e.g., minimum/maximum area and compactness),
    so that elongated or irregular regions inconsistent with a berry silhouette are rejected.

    \item \textbf{Centroid Estimation.} 
    The final 2D reference point $(c_x,c_y)$ is computed as the centroid of the dominant
    retained contour using image moments, and then mapped back to the original image
    coordinates by adding the ROI offset.

    \item \textbf{Fallback Strategy.} 
    If colour segmentation fails within the ROI (e.g., due to specular reflections or low
    saturation), the method falls back to the geometric centre of the YOLO bounding box to
    ensure robustness and avoid tracking loss.
\end{enumerate}

This cascaded design leverages the detector to determine a reliable region of interest
(ROI), and uses HSV-based segmentation to refine where the visible fruit mass lies within
that region.
As a result, the estimated 2D point is typically better aligned with the true fruit region
than the bounding-box centre alone, while maintaining a reliable fallback when local
colour cues are unreliable.

\begin{figure}[h]
\centering

\begin{subfigure}[t]{0.24\textwidth}
    \centering
    \includegraphics[width=\linewidth]{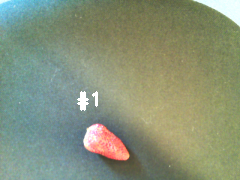}
    \caption{RGB Input}
    \label{fig:rgb}
\end{subfigure}
\hfill
\begin{subfigure}[t]{0.24\textwidth}
    \centering
    \includegraphics[width=\linewidth]{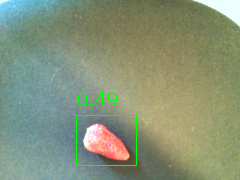}
    \caption{YOLO Detection}
    \label{fig:yolo}
\end{subfigure}
\hfill
\begin{subfigure}[t]{0.24\textwidth}
    \centering
    \includegraphics[width=\linewidth]{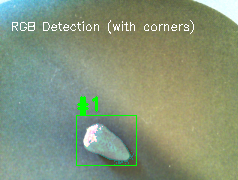}
    \caption{HSV Mask}
    \label{fig:hsv_mask}
\end{subfigure}
\hfill
\begin{subfigure}[t]{0.24\textwidth}
    \centering
    \includegraphics[width=\linewidth]{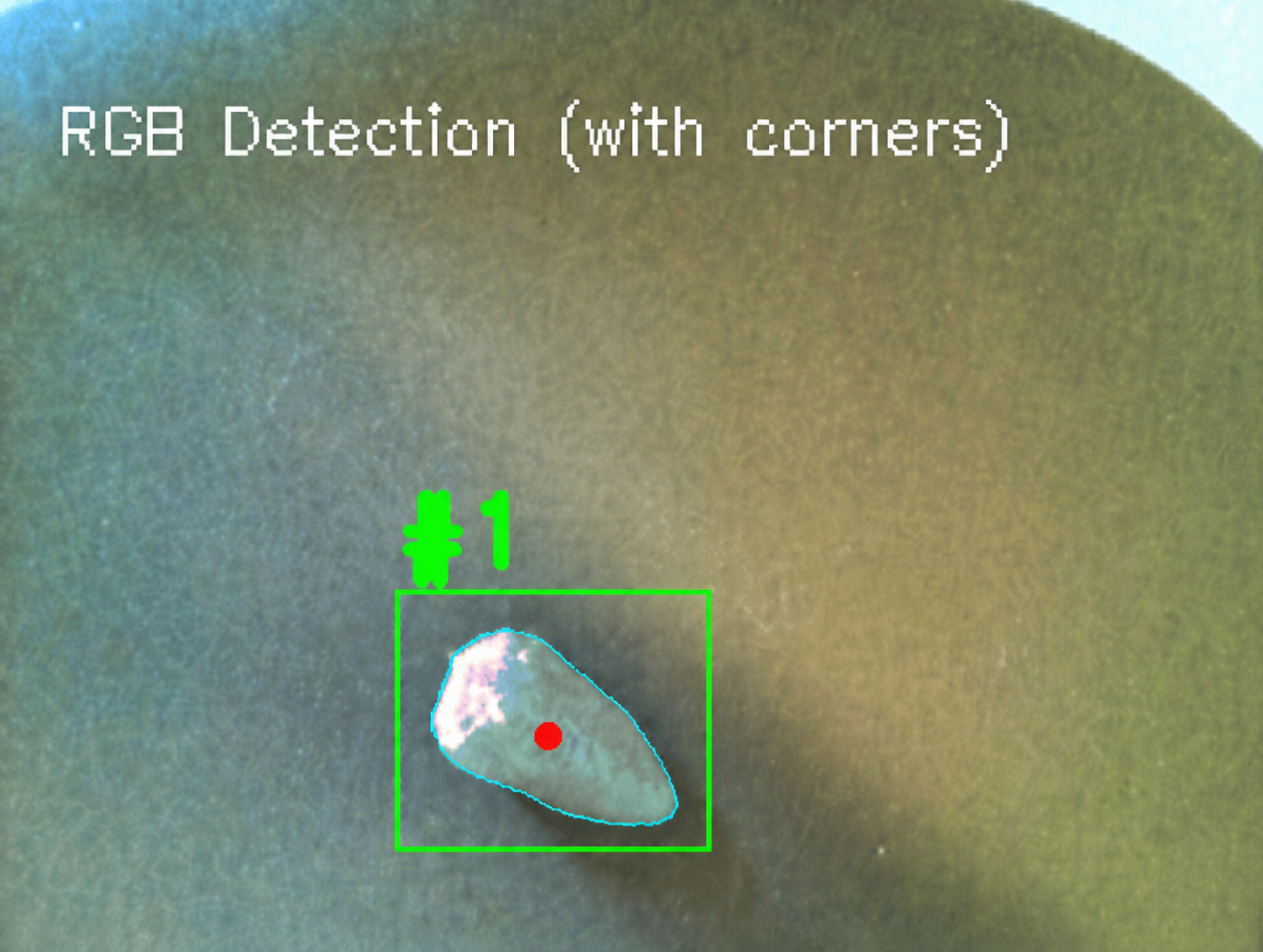}
    \caption{Centroid Extraction}
    \label{fig:centroid}
\end{subfigure}

\caption{Pipeline of 2D strawberry localization: (a) RGB input image, (b) YOLO-based object detection, (c) HSV-based segmentation mask, and (d) centroid extraction for 2D reference point estimation.}
\label{fig:localization_pipeline}

\end{figure}
\paragraph{Linking 2D detections to ToF depth}
\label{sec:linking_2d_tof_depth}
At runtime the strawberry appears as a 2D cue in the RGB image (centroid and, when available,
a bounding box) produced by the detector described earlier.
Range measurements arrive on the ToF depth map defined on the ToF sensor grid, which is not
pixel-aligned with the RGB image.
The linking stage therefore selects a consistent subset of ToF depth samples that geometrically
agree with the 2D detection, and converts the result into a 3D point in the RGB camera frame.

The geometric relationship between the two views is taken from the offline RGB--ToF calibration
stored in \texttt{calib\_results.npz} (Sec.~\ref{sec:rgb_tof_calib}): intrinsics and distortion
for both modalities, together with the rigid transform from the ToF frame to the RGB frame.

Operationally, each valid ToF depth pixel is lifted to a 3D point in the ToF frame, transformed
into the RGB frame, and projected into the RGB image plane so that every ToF measurement carries
an associated image coordinate in the \emph{calibration} resolution.
The live detection is mapped into that same pixel space (including undistortion of the centroid
when distortion coefficients are available) before any distance-based selection is applied.

\begin{enumerate}
    \item \textbf{Nearest-projection linking}
    This policy returns the ToF-derived 3D point whose projection minimises the Euclidean distance to
the undistorted centroid in the calibration image.
It is lightweight, but it can be sensitive to noise and accidental alignment with background
structure when a spurious projection happens to lie close to the centroid.
Matches beyond a fixed pixel-distance threshold are rejected.
\item \textbf{Bounding-box constrained median linking}
This policy retains only those ToF measurements whose projections fall inside the detector
bounding box after the same resolution mapping.
A representative depth is estimated from this subset using a robust statistic with a
foreground-oriented depth window, which reduces bias when the box also contains distant clutter.
Lateral coordinates are then reconstructed along the centroid viewing ray at the estimated depth.
If too few samples fall inside the box, the implementation falls back to~(1).
\end{enumerate}

\paragraph{Depth Sampling Strategy.}
Reliable depth estimation is critical for stable 3D localisation and downstream robotic approach. In practice, pixels within the region of interest (ROI) often contain mixed measurements, including valid strawberry surface points, background leakage, and ToF artefacts such as missing values, spikes, and multipath reflections. Therefore, the depth sampling strategy is designed to preserve the nearest valid surface while mitigating sensitivity to noisy measurements.

\paragraph{Foreground-Cluster Depth Selection.}
The primary approach is based on foreground-cluster selection within the detected ROI. Instead of directly aggregating all ROI depth values, a small fraction (5\%) of extreme near-depth outliers is first removed. A bounded depth interval is then selected, anchored at the minimum reliable depth:

\[
\mathcal{F}=\{z \mid Z_{\min}\le z \le Z_{\min}+120\text{ mm}\}
\]

This formulation is motivated by the scene prior that the strawberry surface is typically located in front of surrounding leaves, stems, and background structures. To ensure robustness, a minimum number of valid samples is required to form a reliable foreground estimate, preventing sparse or unstable clusters from dominating the result.

A trade-off exists between foreground coverage and background suppression. A narrow depth window improves background rejection but may discard valid surface points under noise or pose variation, whereas a wider window increases robustness at the cost of potential background inclusion. Empirically, a window size of 120 mm provides a stable balance at the typical sensing distances used in this work.

\paragraph{Fallback Patch-Based Sampling.}
Foreground-cluster selection relies on sufficient and reliable ROI depth support. In cases where the depth distribution is sparse or inconsistent, a fallback strategy based on local patch sampling is employed.

A square neighbourhood centred at the target pixel is used, corresponding to a \textbf{15$\times$15} patch. This approach prioritises robustness over geometric selectivity. Smaller patches (e.g., \(5\times5\)) are more sensitive to noise, while larger patches are more likely to include background structures. The selected patch size provides a practical compromise between noise suppression and spatial locality.

Within the patch, depth values are filtered using validity bounds (30--1000 mm), followed by trimmed statistics to remove extreme values while preserving the central tendency of the distribution. When amplitude information is available, low-confidence measurements are further suppressed based on an amplitude threshold to improve reliability.

\paragraph{Two-Path Design Rationale.}
A single sampling strategy is insufficient to handle the variability of real-world sensing conditions. The foreground-cluster approach provides stronger geometric selectivity and more effective background rejection when ROI depth is reliable, whereas the patch-based fallback offers improved robustness under degraded sensing conditions. The combination of these two strategies enables both accuracy under nominal conditions and stability under adverse conditions, which is essential for closed-loop robotic operation.

\paragraph{Temporal Stabilisation and Practical Impact.}
To further improve consistency, temporal stabilisation is applied using a sliding median filter with a window size of 5 frames. In addition, depth estimates are constrained within the valid camera operating range (0.08--0.95 m).

Experimental observations indicate that this combined strategy significantly reduces depth fluctuations compared to naive single-pixel or full-ROI averaging methods, resulting in a more stable 3D target for downstream robot motion planning. This directly improves the reliability of localisation and the stability of the closed-loop robotic approach.

\paragraph{Parameter Tradeoffs in Practice}
The most influential parameters are summarized in Table~\ref{tab:depth_tradeoff_i}.

\begin{table}[h]
\centering
\caption{Main parameter tradeoffs in depth sampling}
\label{tab:depth_tradeoff_i}
\begin{tabular}{p{3.2cm}p{1.5cm}p{4.3cm}p{4.3cm}}
\hline
\textbf{Parameter} & \textbf{Value} & \textbf{If Too Small} & \textbf{If Too Large} \\
\hline
Foreground outlier rejection & \(5\%\) & Near spikes remain and destabilize anchoring & True near-surface strawberry points may be removed \\
Foreground depth window & 120 mm & Insufficient foreground support & Increased background leakage \\
Fallback patch size & \(15\times15\) & Sensitive to single-pixel noise and holes & Over-smoothing and background mixing \\
Trim range & 12--88\% & Weak outlier rejection & Over-trimming useful points \\
\hline
\end{tabular}
\end{table}

\paragraph{Camera-frame 3D representation}
\label{par:camera_frame_3d}
After the depth-linking stage (Sec.~\ref{sec:linking_2d_tof_depth}), the strawberry centroid is
available as a Euclidean point $\mathbf{p}_{\mathrm{cam}}=[X,Y,Z]^{\top}$ expressed in the
\texttt{cam\_front} coordinate frame.
As defined in Sec.~\ref{sec:handeye_calibration}, \texttt{cam\_front} is the ROS frame rigidly
attached to the wrist-mounted RGB--ToF assembly; it is published as a child of
\texttt{gripper\_link} with a fixed transform obtained from hand--eye calibration.

Unless otherwise noted, $(X,Y,Z)$ follow the usual \emph{optical} convention used in computer
vision and in the perception nodes: $Z$ points forward along the viewing direction, $X$ to the
right in the image, and $Y$ downward.
If the published URDF or camera driver instead treats \texttt{cam\_front} as a robot link frame
with $x$ forward, a constant axis permutation is applied at runtime so that all subsequent
quantities are expressed in the optical convention before querying TF\@.

At this point $\mathbf{p}_{\mathrm{cam}}$ is still \emph{not} expressed in the manipulator
base; it moves with the arm only indirectly through the time-varying
\texttt{base\_link}$\rightarrow$\texttt{gripper\_link} kinematics, while remaining constant
in \texttt{cam\_front} for a fixed world point whenever the camera--gripper extrinsic is
accurate.
\paragraph{Robot-frame localization via TF}
The camera-frame point $\mathbf{p}_{\mathrm{cam}}$ is published as a stamped 3D point with
\texttt{frame\_id\,=\,cam\_front} and transformed into the robot using the ROS~2 transform
library (\texttt{tf2}).
The transform chain from \texttt{cam\_front} to \texttt{base\_link} encodes the time-varying
hand--eye geometry (camera mounted on the moving end-effector) together with the articulated
robot kinematics, which are maintained as a tree of stamped transforms on \texttt{/tf}.
The transformed point
\[
  \mathbf{p}_{\mathrm{base}} = {}^{\mathrm{base}}\mathbf{T}_{\mathrm{cam}}(t)\,
  \widetilde{\mathbf{p}}_{\mathrm{cam}},
\]
where $\widetilde{\mathbf{p}}_{\mathrm{cam}}$ denotes homogeneous coordinates,
defines the strawberry centroid in the robot base frame used for motion planning and logging.
If the transform is temporarily unavailable, the camera-frame point can still be visualised,
but base-frame quantities are not published until TF succeeds.

\subsubsection{Strawberry Approach (Motion Planning)}
\label{sec:motion_planning}

The motion module generates feasible trajectories that move the arm from the current
configuration toward the strawberry.
Planning operates on the Cartesian target already expressed in \texttt{base\_link}
(Sec.~\ref{sec:localization_base}), so perception and control are connected only through
that 3D point (and optional orientation cues), without re-solving the vision pipeline
inside the planner.

A hybrid pipeline is used: a fast numerical inverse-kinematics (IK) stage proposes
joint goals, and MoveIt~2 plans and executes smooth, collision-aware joint-space trajectories
between successive goals.

\subsection{Robotic Search and Approach (Planning and Control Layer)}
\subsubsection{Strawberry Search}

To locate strawberries efficiently, a waypoint-based scanning strategy is adopted. Instead of performing exhaustive exploration, the robotic arm is controlled to move between a small number of predefined joint configurations, enabling systematic coverage of the target region.

\paragraph{Assumption}

This approach is based on the assumption that strawberries are located in front of the robotic arm within a constrained workspace. This assumption is valid in structured environments such as tabletop experiments or greenhouse rows, where fruits are typically positioned within a predictable frontal region relative to the robot base.

\paragraph{Waypoint Configuration}

The waypoints are defined directly in joint space as a set of predefined joint angle configurations:
\begin{verbatim}
_SEARCH_DEG = [
    (8, -2, -15, 97, -89),
    (8, -2, -143, 143, -89),
]
\end{verbatim}

Each tuple represents the angular positions (in degrees) of the five joints of the robotic arm. The first configuration corresponds to a higher viewing angle, while the second configuration tilts the arm downward, enabling the camera to observe the lower region of the workspace.
\begin{figure}[htbp]
    \centering
    \begin{subfigure}{0.48\textwidth}
        \centering
        \includegraphics[width=\linewidth]{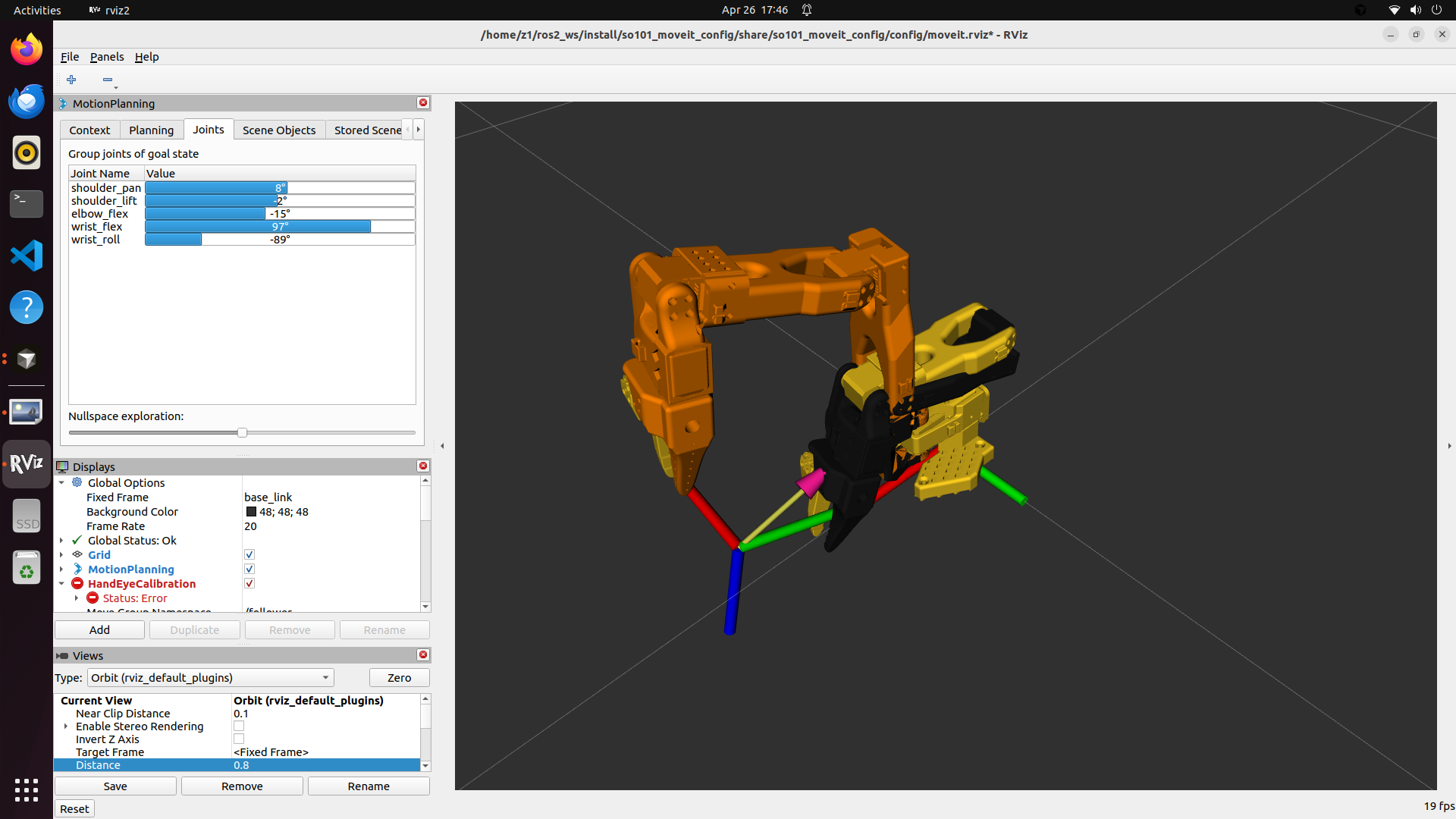}
        \caption{Waypoint 1 (8, -2, -15, 97, -89)}
        \label{fig:waypoint1}
    \end{subfigure}
    \hfill
    \begin{subfigure}{0.48\textwidth}
        \centering
        \includegraphics[width=\linewidth]{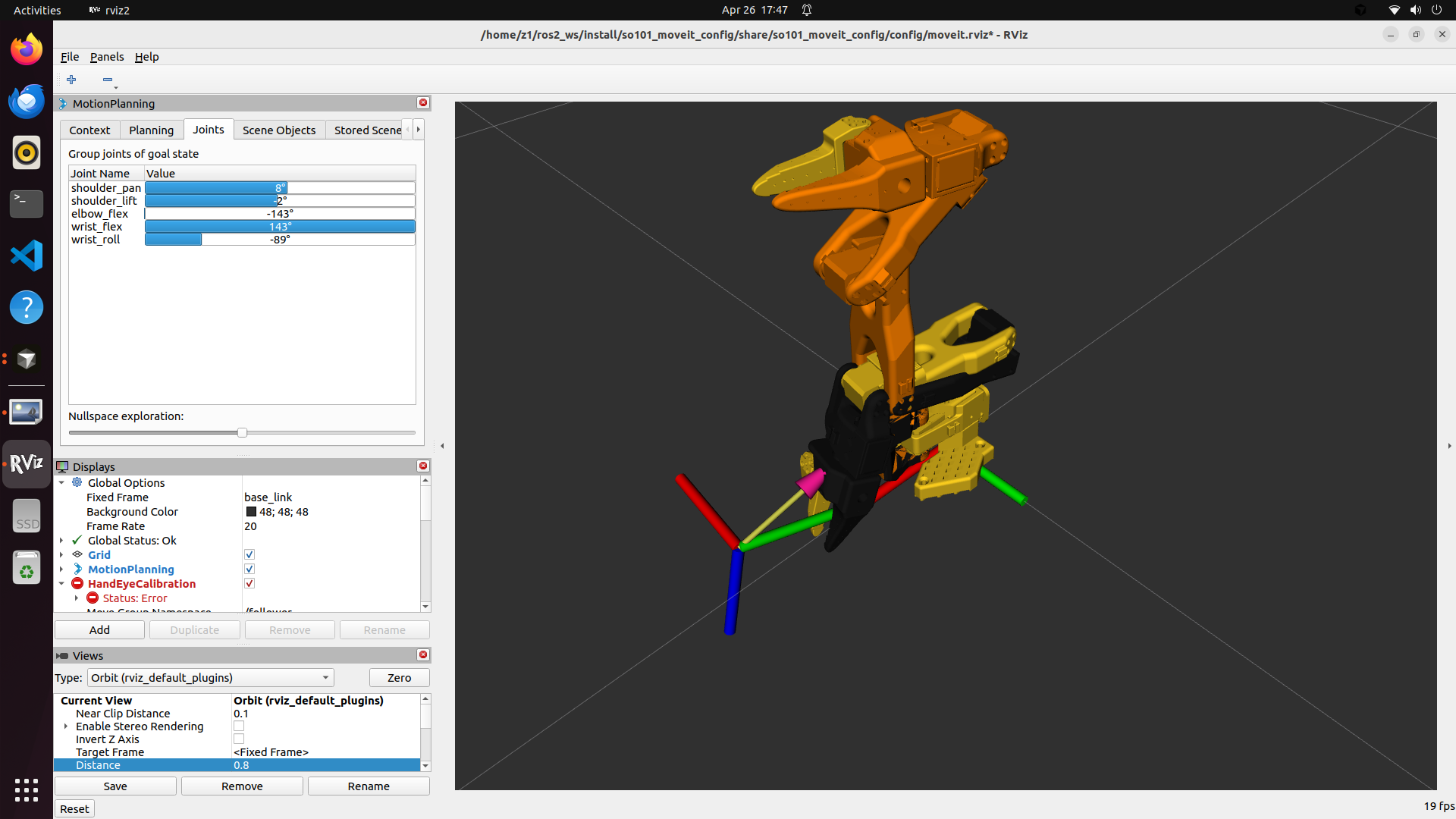}
        \caption{Waypoint 2 (8, -2, -143, 143, -89)}
        \label{fig:waypoint2}
    \end{subfigure}
    
    \caption{Waypoint Configuration}
    \label{fig:two_images}
\end{figure}
\paragraph{Search Strategy}

The robotic arm alternates between these two waypoints to perform a vertical scanning motion. Specifically:
\begin{enumerate}
    \item The arm moves to the upper waypoint to observe the upper portion of the scene
    \item It then moves to the lower waypoint to capture the lower region
    \item At each waypoint, RGB and ToF data are captured and processed for strawberry detection
\end{enumerate}

This motion effectively creates an up-and-down scanning pattern, ensuring coverage of the frontal workspace without requiring complex trajectory planning.

\paragraph{Design Rationale}

The waypoint design focuses on simplicity and efficiency:
\begin{itemize}
    \item \textbf{Vertical coverage:} By varying the third and fourth joint angles, the camera’s viewing direction changes significantly, enabling coverage of different heights
    \item \textbf{Fixed lateral position:} The first and second joint angles remain nearly constant, ensuring that the camera consistently faces forward
\end{itemize}

This design allows the robot to quickly scan the most relevant region where strawberries are expected to appear.

\paragraph{Integration with Detection and Control}

Once a strawberry is detected at any waypoint, the scanning process terminates. The detected position is then used as input to the downstream pipeline, where hand-eye calibration and inverse kinematics are applied to guide the robotic arm towards the target for close-range sensing. A closed-loop refinement strategy is subsequently employed to improve positioning accuracy.

\paragraph{Numerical inverse kinematics (IK)}
For a desired end-effector pose (\texttt{gripper\_link}), joint angles are found with a constrained optimisation
solver (L-BFGS-B in the implementation).
The cost penalises Cartesian position error relative to the target while respecting joint
limits and a floor height so the gripper does not violate workspace safety bounds.
A small regularisation term on joint displacement relative to the current configuration
prefers smooth, incremental posture changes instead of large jumps between IK solutions.

\paragraph{IK objective (loss).}
Let $\mathbf{q}\in\mathbb{R}^5$ denote the arm joints and let $\mathbf{p}_e(\mathbf{q})$ be the
Cartesian position of \texttt{gripper\_link} in \texttt{base\_link}, obtained from the
same forward kinematics chain as in calibration.
Given a Cartesian target $\mathbf{t}$ and the current configuration $\mathbf{q}_0$ used for
regularisation, the IK solver minimises the scalar cost
\begin{equation}
  J(\mathbf{q}) =
  \underbrace{\|\mathbf{p}_e(\mathbf{q})-\mathbf{t}\|_2^2}_{\text{position}}
  \;+\;
  \underbrace{w_z\,\bigl[\max\bigl(0,\,z_{\mathrm{floor}}-p_{e,z}(\mathbf{q})\bigr)\bigr]^2}_{\text{height floor}}
  \;+\;
  \underbrace{w_{\mathrm{reg}}\,\|\mathbf{q}-\mathbf{q}_0\|_2^2}_{\text{joint regularisation}} .
  \label{eq:ik_loss_base}
\end{equation}
The height term softly enforces a minimum gripper height $z_{\mathrm{floor}}$ in
\texttt{base\_link}; it is inactive when $p_{e,z}\ge z_{\mathrm{floor}}$.
The regularisation term discourages large joint jumps between successive IK solves along the
incremental approach.

\paragraph{Camera--target alignment}
A mild constraint is introduced to keep the camera roughly facing the strawberry during the approach.
This helps maintain the target near the image centre, improving the stability of repeated closed-loop updates.

When a look-at target $\boldsymbol{\ell}$ is provided (the strawberry position in
\texttt{base\_link}), an additional \emph{viewing} term biases the camera optical axis toward
the fruit.
Let $\mathbf{c}(\mathbf{q})$ and $\mathbf{z}_c(\mathbf{q})$ be the camera position and optical
axis (unit vector) in \texttt{base\_link}, and let
$\mathbf{d}(\mathbf{q})=(\boldsymbol{\ell}-\mathbf{c})/\|\boldsymbol{\ell}-\mathbf{c}\|$
(with the term skipped if $\|\boldsymbol{\ell}-\mathbf{c}\|$ is negligible).
The full cost used in code is
\begin{equation}
  J_{\mathrm{full}}(\mathbf{q}) =
  J(\mathbf{q})
  \;+\;
  w_{\mathrm{view}}\Bigl(1-\mathbf{z}_c(\mathbf{q})^{\top}\mathbf{d}(\mathbf{q})\Bigr),
  \label{eq:ik_loss_lookat}
\end{equation}
where $1-\mathbf{z}_c^{\top}\mathbf{d}$ is small when the camera faces the target and increases
as the optical axis deviates from the line of sight.

Equation~\eqref{eq:ik_loss_lookat} is minimised with SciPy's bounded quasi-Newton routine
(\texttt{L-BFGS-B}) subject to joint limits, using the current $\mathbf{q}_0$ and several
randomly perturbed seeds to reduce the risk of local minima.
A candidate is accepted only if $\|\mathbf{p}_e(\mathbf{q})-\mathbf{t}\|$ is below a fixed
positional tolerance after optimisation.

\paragraph{Trajectory generation with MoveIt~2}
Once an IK solution is accepted (position error below a fixed tolerance), the target
configuration is sent to MoveIt~2's \texttt{move\_group} interface, which plans a
collision-free trajectory in joint space and executes it on the hardware controller.
Internally, the planner is sampling-based (e.g., RRT-family planners, depending on the
MoveIt configuration), subject to the robot model and scene collision geometry.

\begin{figure}[H]
    \centering
    \includegraphics[width=0.95\linewidth]{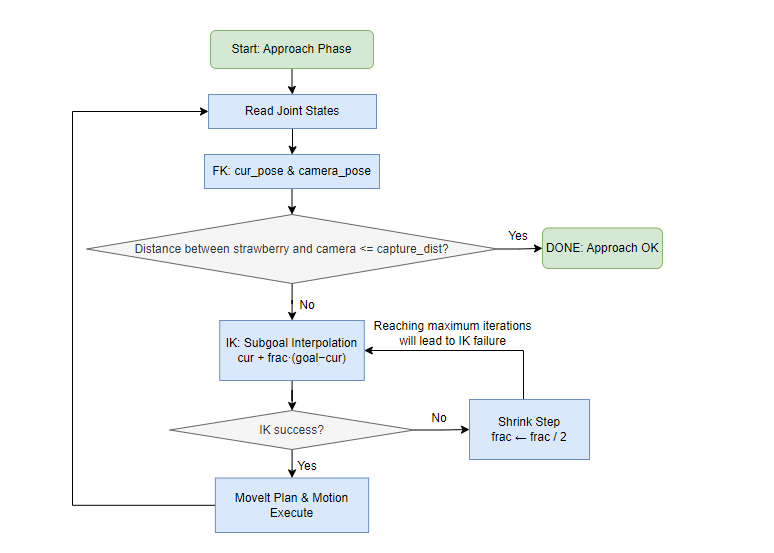}
    \caption{Closed-loop perception and control pipeline.}
    \label{fig:closed_loop}
\end{figure}

\paragraph{Closed-loop Approach Strategy}
The robotic approach phase is implemented as an iterative closed-loop control process, as illustrated in Figure~\ref{fig:closed_loop}. 
The objective is to progressively move the end-effector towards the target strawberry while ensuring stability and robustness against perception and kinematic uncertainties.

The process begins by reading the current joint states of the robotic arm, followed by FK to compute the current end-effector pose as well as the camera pose mounted on the wrist. 
Based on the estimated 3D position of the strawberry, the system evaluates the distance between the camera and the target.

If the distance is below a predefined capture threshold, the approach phase terminates successfully. 
Otherwise, the system proceeds to generate an incremental motion towards the goal using an IK-based interpolation strategy. 
Specifically, the next target configuration is computed as a fractional step between the current pose and the desired goal pose.

The IK solver attempts to find a feasible joint configuration for this interpolated target. 
If the IK solution is successful, the system executes the motion using the motion planning module, and the loop restarts with updated joint states. 
If the IK solver fails, the step size is reduced by shrinking the interpolation factor, effectively taking a smaller step towards the goal to improve feasibility.

This adaptive step-size strategy continues iteratively until a valid IK solution is found or a maximum number of iterations is reached, in which case the approach is considered unsuccessful. 
By continuously updating the robot state and refining the motion commands, the system achieves a robust closed-loop behavior that mitigates the effects of kinematic constraints and environmental uncertainty.
\subsection{Non-Destructive Sweetness Measurement}
\subsubsection{Experimental Setup}
All experiments were conducted in a windowless room to eliminate external light interference and maintain stable temperature, reducing measurement variability and ensuring data consistency. A custom fixture securely held and rotated each strawberry for consistent multi-angle acquisition.

\begin{figure}[htbp]
    \centering
    \begin{subfigure}{0.4\textwidth}
        \centering
        \includegraphics[width=\linewidth]{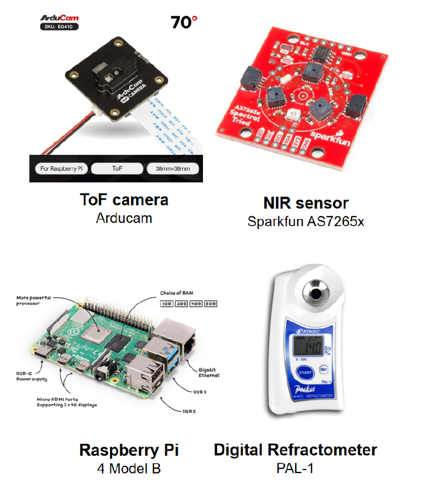}
        \caption{Hardware Devices}
        \label{fig:devices}
    \end{subfigure}
    \hfill
    \begin{subfigure}{0.48\textwidth}
        \centering
        \includegraphics[width=\linewidth]{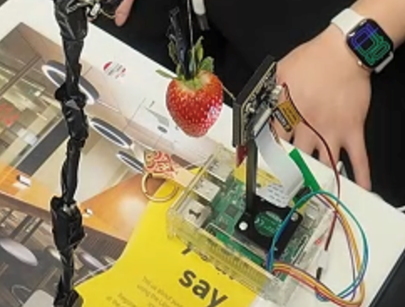}
        \caption{Experimental Setup}
        \label{fig:devices2}
    \end{subfigure}
    \caption{Overview of the experimental setup.}
    \label{fig:overall}
\end{figure}

The main devices \ref{fig:devices} included:

\begin{itemize}
    \item \textbf{Arducam ToF camera:} Captured high-resolution depth images for spatial and morphological information.
    
    \item \textbf{Raspberry Pi 4 Model:} Used for data acquisition, storage, and model inference.
    
    \item \textbf{PAL-1 digital refractometer:} Measured average Brix values as ground truth for the classification model.
\end{itemize}

The combination of a controlled environment and high-performance devices ensured the accuracy and repeatability of data for both the ToF classification.

\subsubsection{Data Collection}
Strawberry samples were acquired in four acquisition rounds, including an initial trial batch (Batch~0) and three subsequent formal batches. 
The trial batch was used to validate the sensing pipeline and refine the experimental setup, including sensor placement, distance configuration, and data acquisition parameters. 
Only data collected after this initial validation stage were considered for the main analysis.

\begin{table}[h]
\centering
\caption{Strawberry sample collection batches}
\label{tab:sample_batches}
\begin{tabular}{l c c}
\hline
\textbf{Batch} & \textbf{\# Strawberries} & \textbf{Valid for ToF} \\
\hline
Batch 0 (trial) & 4  & 3 \\
Batch 1         & 4  & 4 \\
Batch 2         & 17 & 17 \\
Batch 3         & 21 & 20 \\
\hline
\textbf{Total} & \textbf{44} & \textbf{40} \\
\hline
\end{tabular}
\end{table}

For each strawberry, close-range spectral measurements were collected using the ToF-based sensing module at a controlled distance (15 and 20~cm). 
During acquisition, multiple frames were captured to account for sensor noise and variability, and signal aggregation (e.g., averaging or median filtering) was applied to improve robustness.

Not all collected samples were considered valid for ToF-based analysis. Invalid samples were excluded due to factors such as unstable depth readings, insufficient signal strength, occlusion, or misalignment between the sensing direction and the fruit surface. 
As shown in Table~\ref{tab:sample_batches}, a total of 40 out of 44 samples were retained for subsequent modelling and evaluation.

The final dataset covers a range of strawberry appearances and conditions, providing a representative basis for evaluating the proposed non-destructive sweetness prediction pipeline.

\subsubsection{Brix distribution by brand}
To achieve a more balanced distribution of strawberry sweetness, I deliberately purchased multiple varieties of strawberries for testing and analysis.

 \begin{figure}[H]
    \centering
    \includegraphics[width=0.75\linewidth]{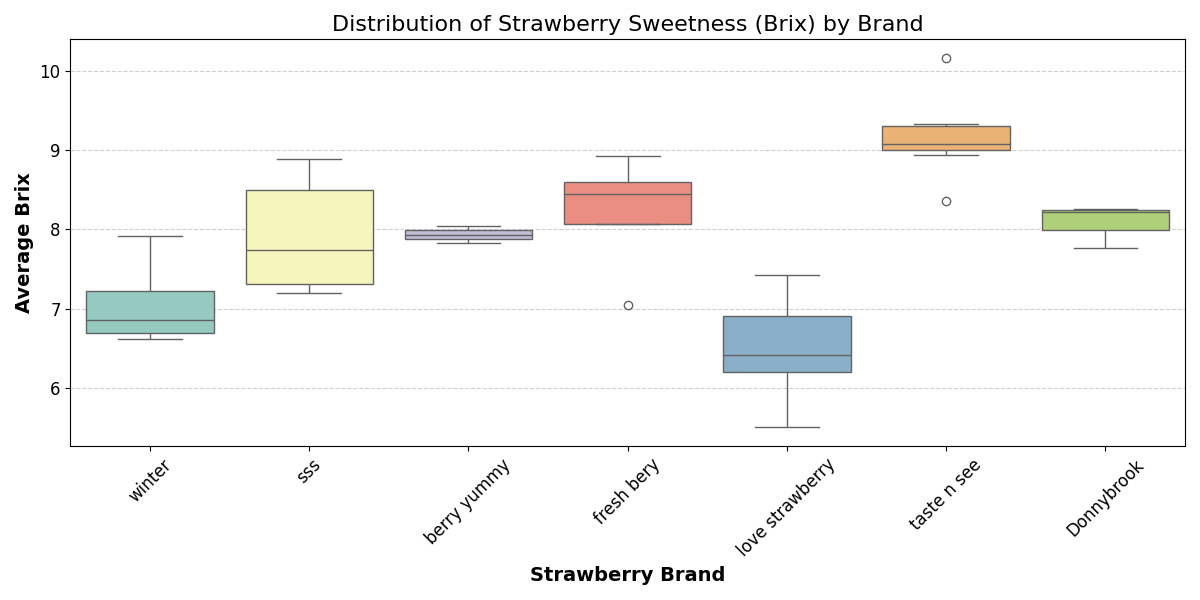}
    \caption{Brix distribution by brand}
    \label{fig:brand_distribution}
\end{figure}

\subsubsection{Time-of-Flight (ToF) Data  }
\begin{figure}[H]
    \centering
    \includegraphics[width=0.7\linewidth]{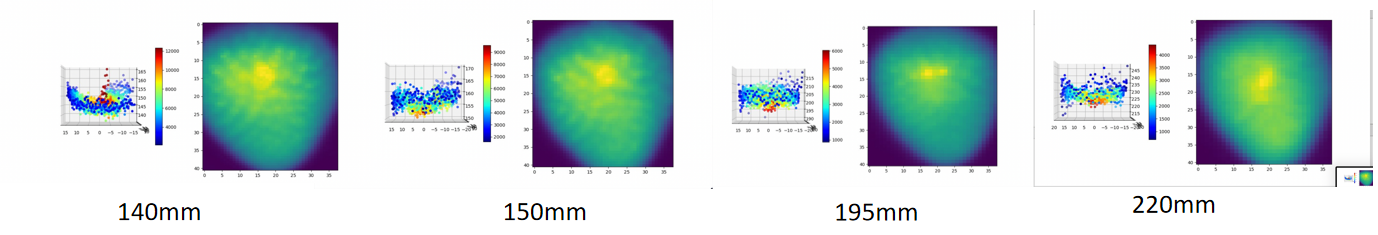}
    \caption{Optimal measurement distance for ToF data acquisition.}
    \label{fig:optimal_distance}
\end{figure}

Time-of-Flight (ToF) measurements were obtained using sensor configurations at 16 cm and 20 cm.
 In each configuration, the strawberry was rotated while capturing 4,000 ToF images.
 Distances ranging from 140 mm to 220 mm were evaluated to investigate the effect of target distance on measurement accuracy.
 For distances shorter than 150 mm, the recorded values were consistently higher—approximately 165 mm—due to internal light scattering within the fruit.
 For distances exceeding 200 mm, the measurements exhibited reduced resolution. \ref{fig:optimal_distance}
 
\subsubsection{Ground-Truth Measurement  }
For the ground truth acquisition, each strawberry was weighed and then juiced.
The juice was passed through multiple layers of filtration using a fine mesh to remove solids.
After obtaining the clarified strawberry juice, both the ambient room temperature and the juice temperature were recorded.
Soluble solids content (°Brix) was then measured seven times using a digital refractometer, and the average value was taken as the reference measurement. 

\subsubsection{Classification Model Approach}
For classification, ToF features were used to train a \textbf{Simple Convolutional Neural Network (SimpleCNN)} developed using PyTorch. The classifier is a compact 3D CNN that maps the 2-channel voxel grid to binary logits (high vs. low sugar). The backbone comprises four convolutional blocks with \textbf{batch normalization} and \textbf{ReLU activations}. Spatial resolution is reduced three times via 2×2 max-pooling; the final block keeps spatial resolution, followed by global average pooling and two fully connected layers, resulting in a binary output as prediction. Training the model took approximately 50 minutes on a \textbf{NVIDIA H200 graphics accelerator} with the voxel grid size of 32, while the model with the grid size of 128 took approximately 12 hours to train. The breakdown of the model is as follows:

\begin{figure}[H]
    \centering
    \includegraphics[width=0.95\linewidth]{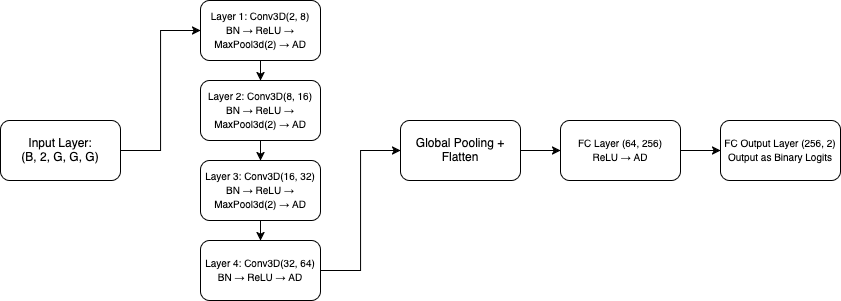}
    \caption{CNN Model Architecture}
    \label{fig:placeholder}
\end{figure}

All dropout layers use an “adaptive” policy where the dropout probability decays linearly from an initial value to a final value over training epochs. This provides stronger regularization early in training and reduces stochasticity as convergence is approached.

Global average pooling is used to replace large fully connected projections on the 3D feature map, reducing parameters and improving robustness to small spatial shifts of the strawberry within the voxel grid.

\subsubsection{Measurement Distance}  
A critical step in optimizing the model involved selecting an appropriate measurement distance. After thorough evaluation, 15\,cm was chosen as the key measurement distance because it strikes a balance between capturing sufficient spatial detail and maintaining operational practicality. At this distance, the sensor can capture meaningful features within the strawberry samples while enabling precise measurement and efficient data processing. Furthermore, the 15\,cm distance aligns well with the spatial resolution of our sensors, ensuring accurate and consistent data acquisition. This choice provided an optimal trade-off between detail and efficiency, which was essential for the success of our analysis.

\subsubsection{Classification Threshold}  
The binary classification threshold for sweetness was set at 8.5\,${}^\circ$Brix based on two main considerations: dataset characteristics and sensory research findings.

\paragraph{Rationale Based on Dataset and Sensory Research}  
Industry guidance suggests that premium greenhouse strawberries typically target approximately 9\,${}^\circ$Brix, which is higher than the basic market standard of 7\,${}^\circ$Brix. Furthermore, sweetness combined with the volatile compound profile plays a crucial role in determining the overall sensory quality of strawberries, providing a scientific rationale for selecting a relatively higher sweetness threshold.

\begin{figure}[H]
    \centering
    \includegraphics[width=0.65\linewidth]{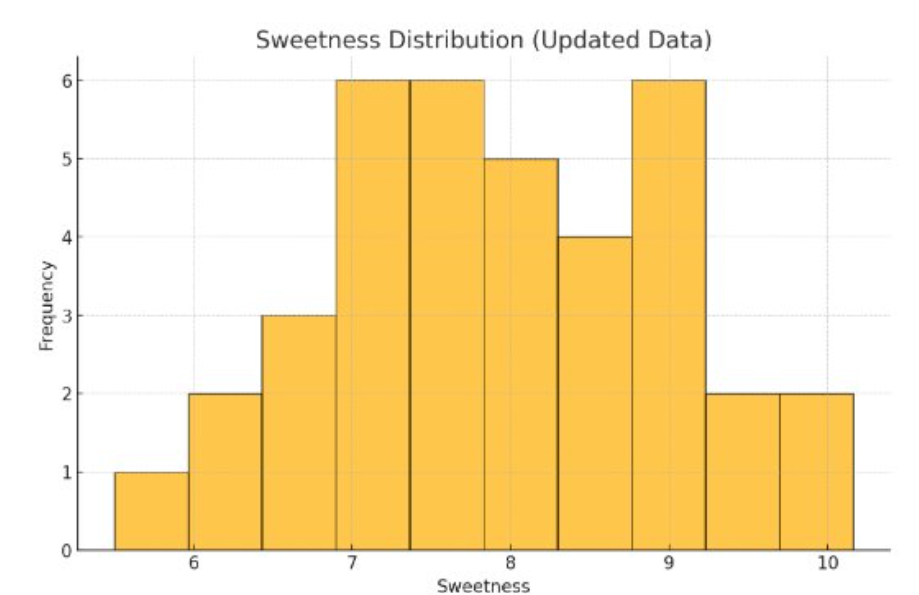}
    \caption{Distribution of strawberry sweetness levels across samples.}
    \label{fig:distribution}
\end{figure}

\paragraph{Rationale Based on Dataset Distribution}  
However, applying a 9\,${}^\circ$Brix cutoff within our dataset would result in too few high-sugar samples, leading to significant class imbalance and potentially reducing classification accuracy. By contrast, selecting a threshold in the range of 7.5 to 8.5\,${}^\circ$Brix offers a more balanced distribution between high- and low-sugar samples, improving model robustness and predictive performance. Based on this, 8.5\,${}^\circ$Brix was chosen as a practical and effective compromise.

It is important to note that, due to the smaller size of strawberries compared to other fruits such as apples, implementing data fusion techniques to enhance classification would require a substantially larger dataset and adjusted model architectures. Given time constraints, fusion was not implemented in this project but remains a key focus for future work.

\subsection{System Integration}
\label{sec:system_integration}

This section summarises how independently developed components are integrated into a unified
ROS~2-based system for non-destructive strawberry sweetness estimation. 
The system combines sensing, perception, motion control, and learning-based prediction modules 
into a coherent runtime pipeline that can be initialised, calibrated once, and repeatedly 
executed for automated data acquisition and sweetness evaluation.

The integrated pipeline enables the robotic platform to autonomously detect strawberries, 
position the sensor at an appropriate distance, acquire close-range spectral measurements, 
and perform real-time sweetness classification. All modules communicate through ROS~2 topics 
and services, ensuring modularity and extensibility while maintaining synchronisation between 
image streams, depth sensing, and robot motion.

Once deployed, the system supports repeatable experiments, 
providing a practical framework for evaluating non-destructive fruit quality assessment in 
realistic scenarios.

\subsubsection{Runtime composition}
The integrated system follows the usual ROS~2 pattern of a robot description (URDF) plus a
bring-up launch file that starts the joint driver, publishes robot state, and exposes the
time-varying transform from \texttt{base\_link} to the wrist.
Static transforms obtained offline attach the RGB--ToF sensor frame \texttt{cam\_front} to
\texttt{gripper\_link} (hand--eye) and, where used, supply the RGB--ToF intrinsics/extrinsics
consumed by the detector.
A workstation-side bridge streams camera data from the embedded platform and republishes it
into the same ROS graph, so perception nodes always subscribe to a consistent topic namespace
regardless of whether the physical link is wired or wireless.

\subsubsection{Perception--planning coupling}
The detector publishes a Cartesian target in \texttt{base\_link}, which the motion stack treats
as an external input to its closed-loop incremental planner.
This loose coupling allows the perception rate, the bridge rate cap, and the planning cycle to
differ: the motion module simply consumes the latest valid target while respecting TF
availability and workspace limits.

\subsubsection{Operational workflow}
In practice, integration is validated by a repeatable start-up sequence: launch base hardware
and TF publishers, start the camera bridge, start perception, then start planning/MoveIt and
the task script.
The same decomposition also supports partial bring-up (e.g., perception-only logging) by
omitting nodes that are not required for the current experiment.

\subsubsection{Reproducibility}
Versioned calibration artefacts (RGB--ToF \texttt{npz}, hand--eye YAML, and optional intrinsics
YAML) are treated as configuration inputs to the launch layer so that different experimental
sessions can be replayed without re-estimating extrinsics, while still allowing A/B comparison
when calibration files are swapped deliberately.
\subsection{Communication pipeline}
\label{sec:communication_pipeline}
\begin{figure}[H]
    \centering
    \includegraphics[width=1\linewidth]{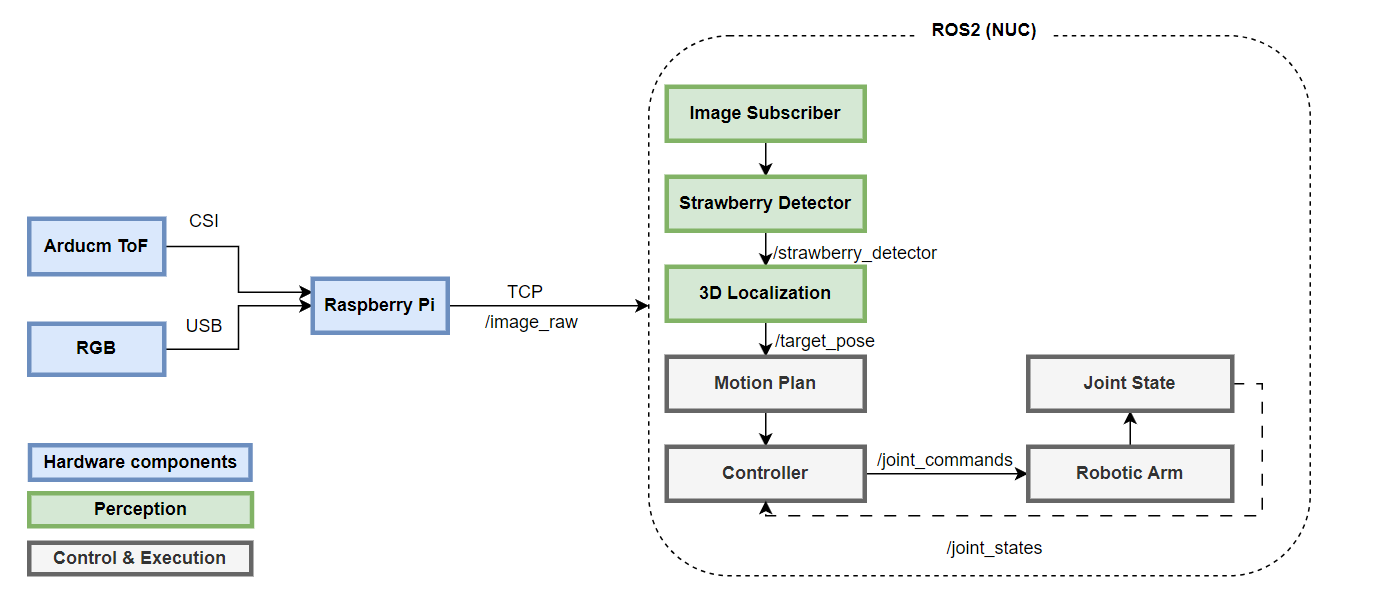}
    \caption{Communication pipeline of the system}
    \label{fig:pipeline}
\end{figure}

Figure~\ref{fig:pipeline} illustrates the overall communication pipeline of the proposed system, which integrates heterogeneous sensing, perception, and robotic control components across a distributed architecture.

On the sensing side, an RGB camera and an Arducam ToF sensor are connected to a Raspberry Pi via USB and CSI interfaces, respectively. The Raspberry Pi acts as an edge device that synchronizes and streams image data to a more powerful computation unit (NUC) over a TCP connection using the \texttt{/image\_raw} topic.

On the NUC, the system is implemented within the ROS2 framework. An image subscriber node first receives the incoming image stream and forwards it to the strawberry detection module. The detector processes the RGB input and outputs candidate regions of interest corresponding to strawberries. These detections are then passed to the 3D localization module, which estimates the spatial position of the target in the robot coordinate frame, publishing the result as a \texttt{/target\_pose}.

The estimated target pose is subsequently used by the motion planning module to generate feasible trajectories for the robotic arm. The planned motion commands are then executed by the controller, which translates high-level plans into low-level joint commands and publishes them via the \texttt{/joint\_commands} topic.

Finally, the robotic arm executes the commands and continuously publishes its joint states through the \texttt{/joint\_states} topic, forming a closed-loop feedback system. This feedback enables the controller to adjust actions in real time, ensuring robust and stable operation during the strawberry approach task.

\section{Results and Discussion}
\subsection{Strawberry Detection and Segmentation}
\subsubsection{Quantitative Results}
\begin{figure}[H]
    \centering
    \begin{subfigure}[b]{0.24\textwidth}
        \centering
        \includegraphics[width=\textwidth]{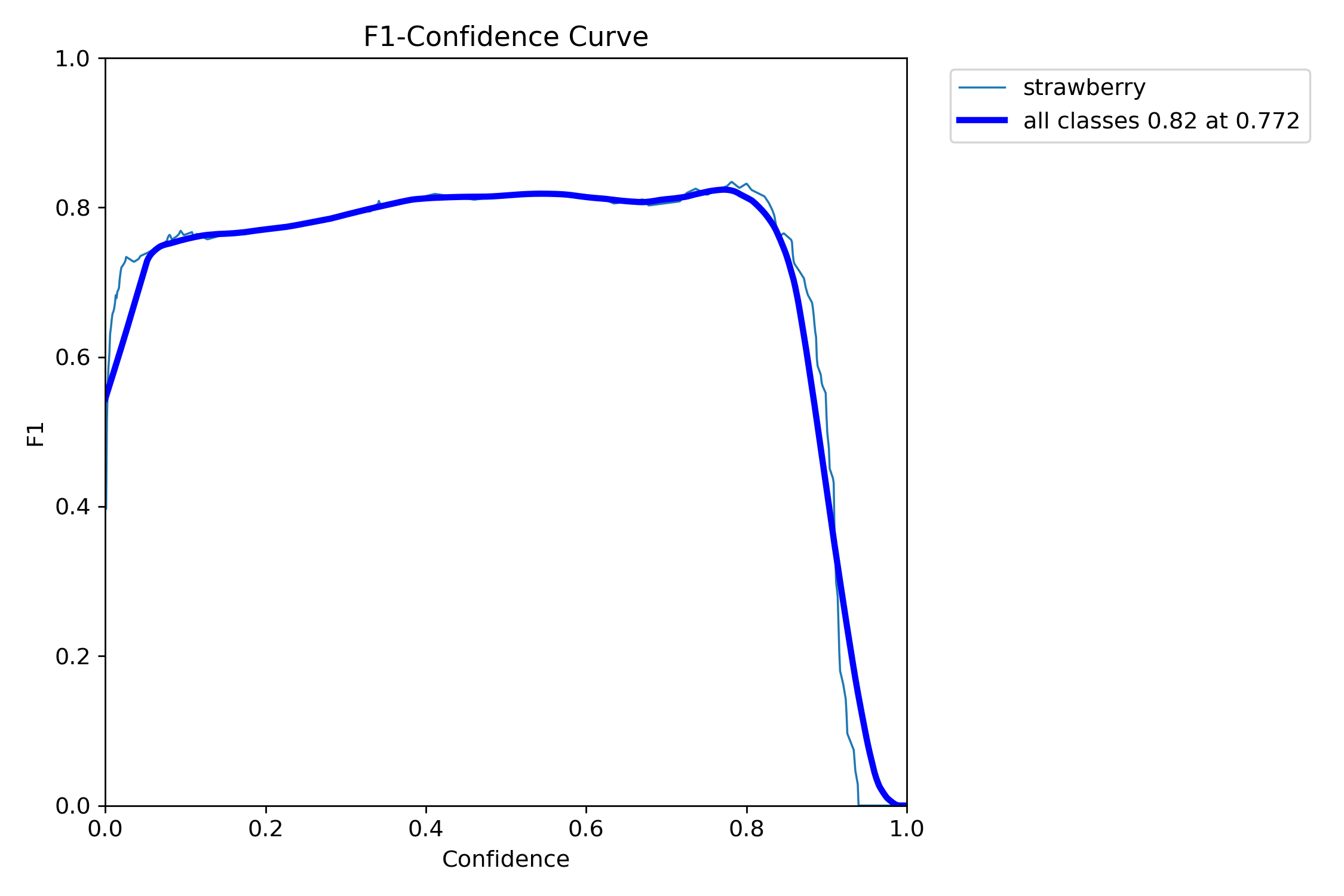}
        \caption{F1 Curve}
        \label{fig:val_single}
    \end{subfigure}
    \hfill
    \begin{subfigure}[b]{0.24\textwidth}
        \centering
        \includegraphics[width=\textwidth]{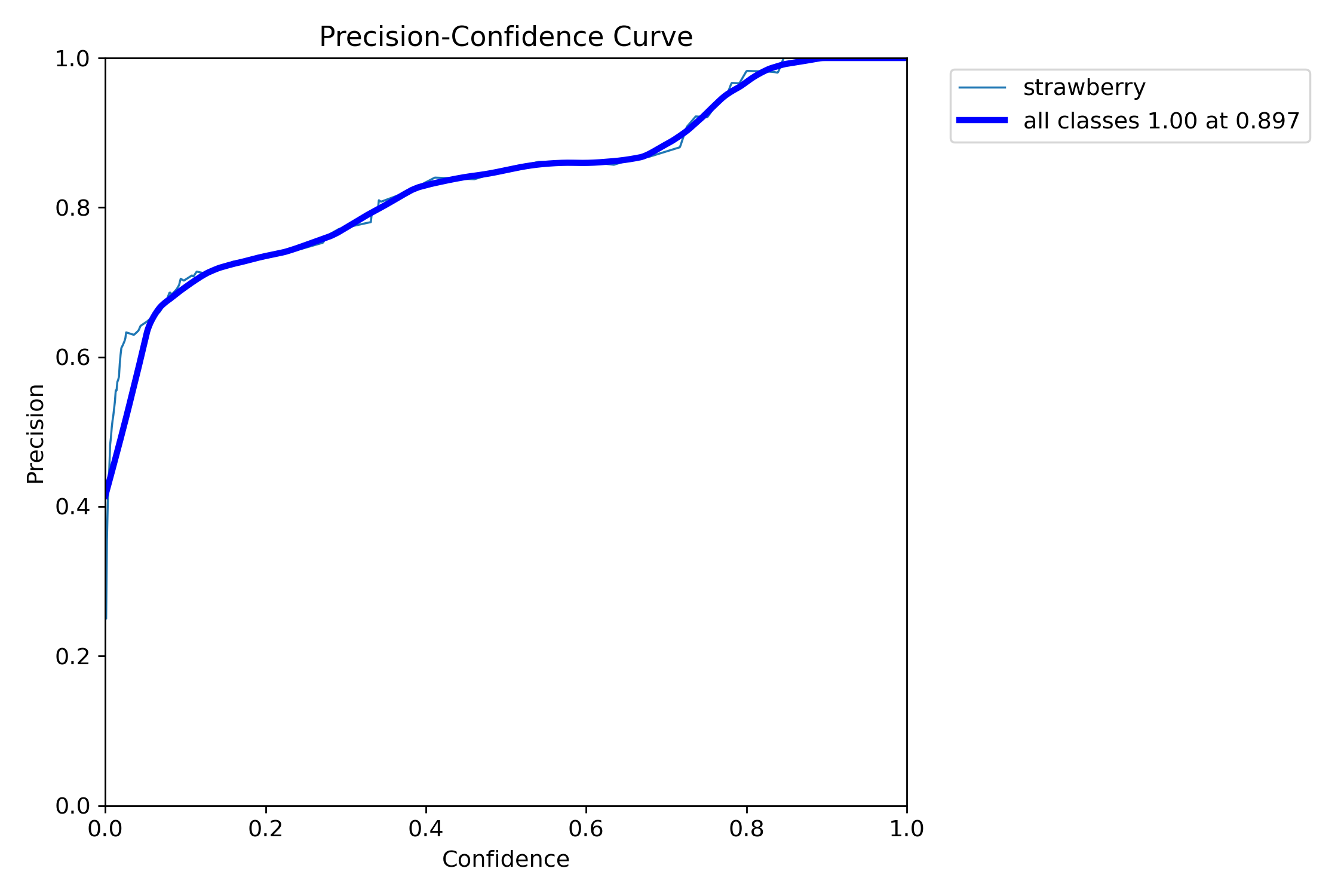}
        \caption{P Curve}
        \label{fig:val_two_p}
    \end{subfigure}
    \hfill
    \begin{subfigure}[b]{0.24\textwidth}
        \centering
        \includegraphics[width=\textwidth]{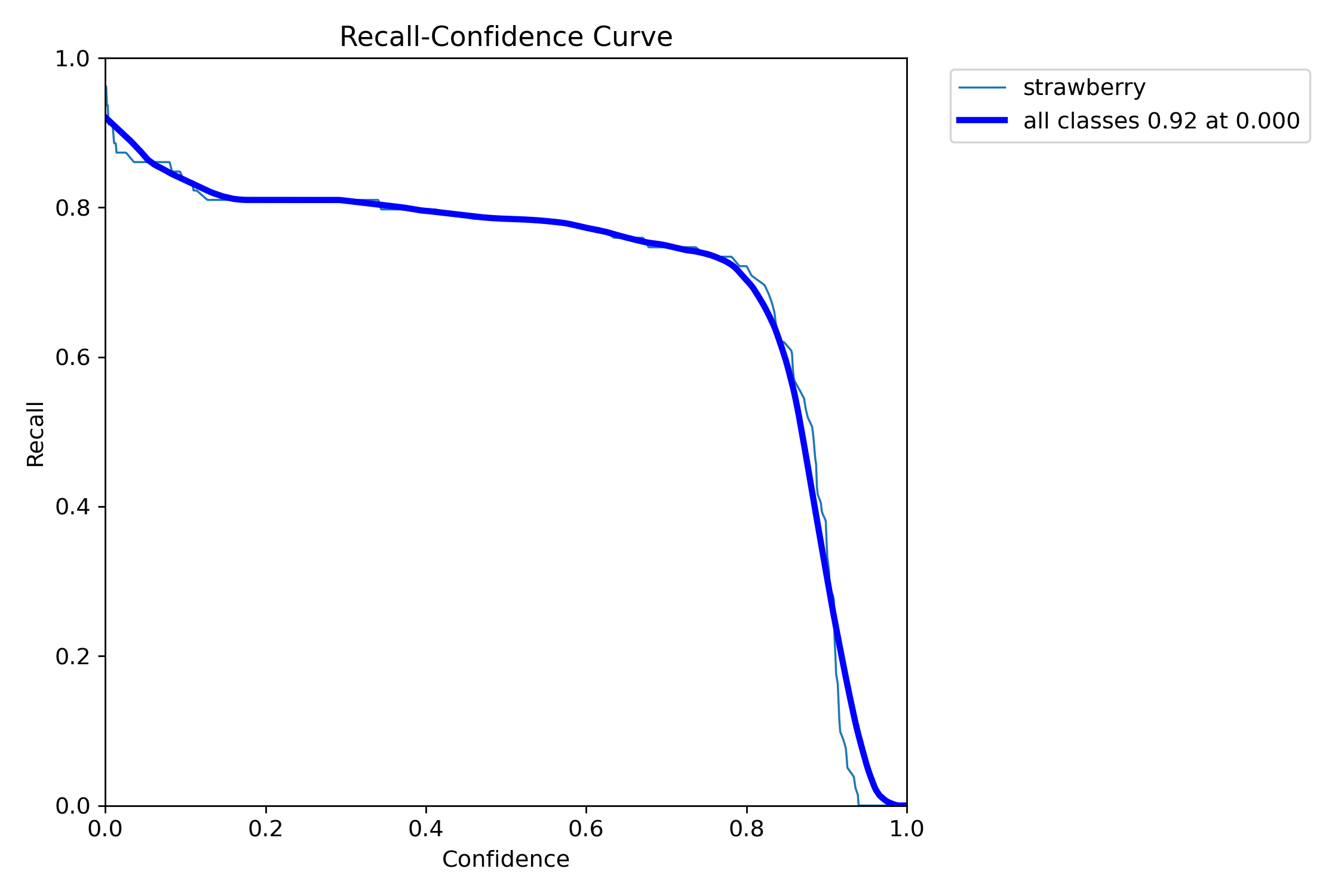}
        \caption{R Curve}
        \label{fig:val_two_r}
    \end{subfigure}
    \hfill
     \begin{subfigure}[b]{0.24\textwidth}
        \centering
        \includegraphics[width=\textwidth]{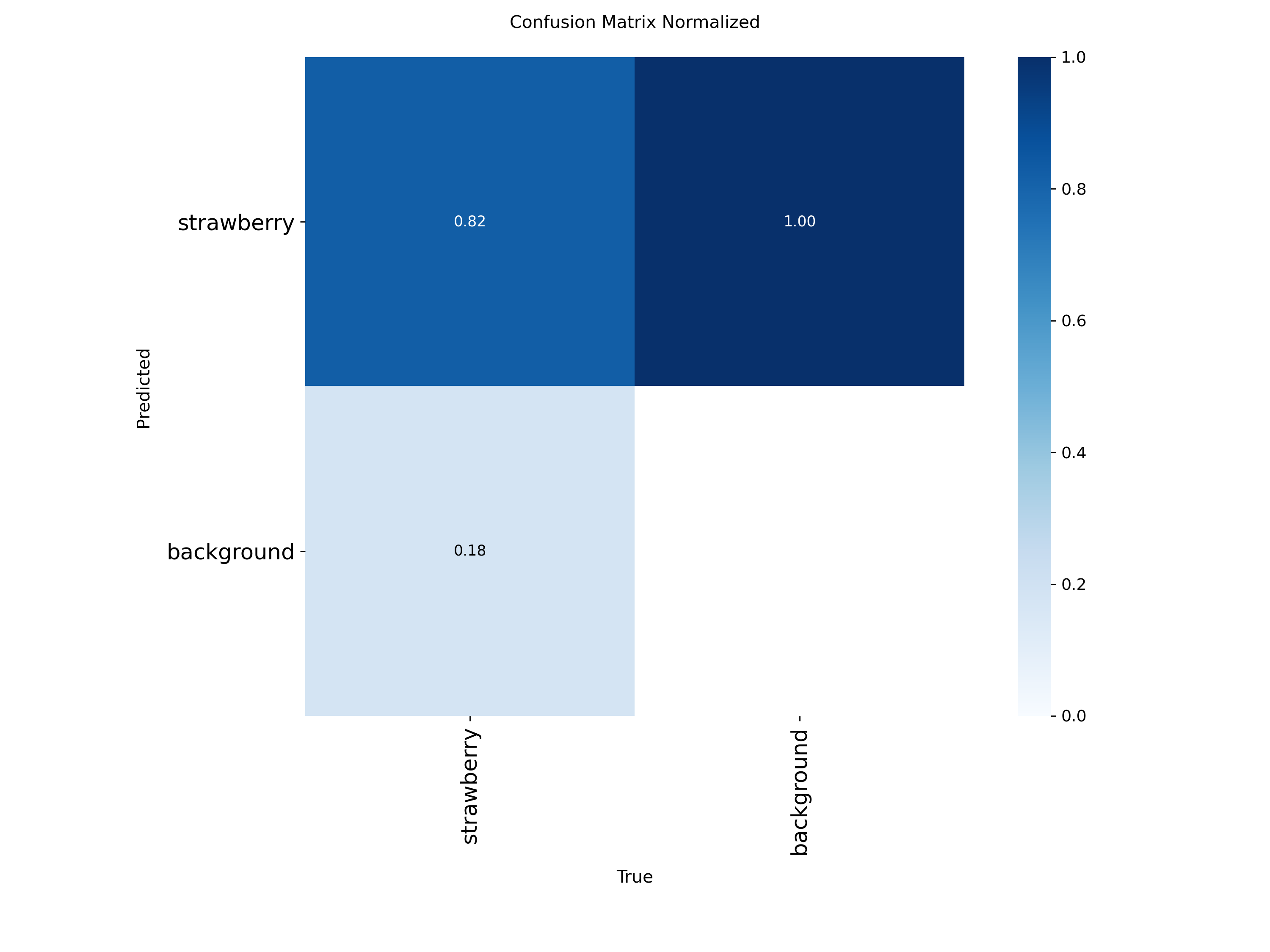}
        \caption{Confusion Matrix}
        \label{fig:confusion_matrix}
    \end{subfigure}

    \caption{Detection validation results using different metrics.}
    \label{strawberry_detection/fig:strawberry_calibration_validation}
\end{figure}

The model demonstrates strong performance in strawberry detection, achieving a peak F1-score \ref{fig:val_single} of 0.82 at a confidence threshold of 0.772, with stable results across a broad confidence range of 0.4 to 0.8. Precision \ref{fig:val_two_p} reaches 1.00 at higher confidence levels (0.897), indicating that the model rarely produces false positives when highly confident. Meanwhile, recall \ref{fig:val_two_r} attains 0.92 at the minimum threshold, showing that the vast majority of strawberries in the dataset are successfully detected. Notably, the normalized confusion matrix \ref{fig:confusion_matrix} reveals that 100\% of actual strawberries are correctly retained, with no false negatives at the class level, and 82\% of detections are precisely classified, further highlighting the reliability and robustness of the model.

\subsubsection{Qualitative Results}
\begin{figure}[H]
    \centering
    \begin{subfigure}[b]{0.48\textwidth}
        \centering
        \includegraphics[width=\textwidth]{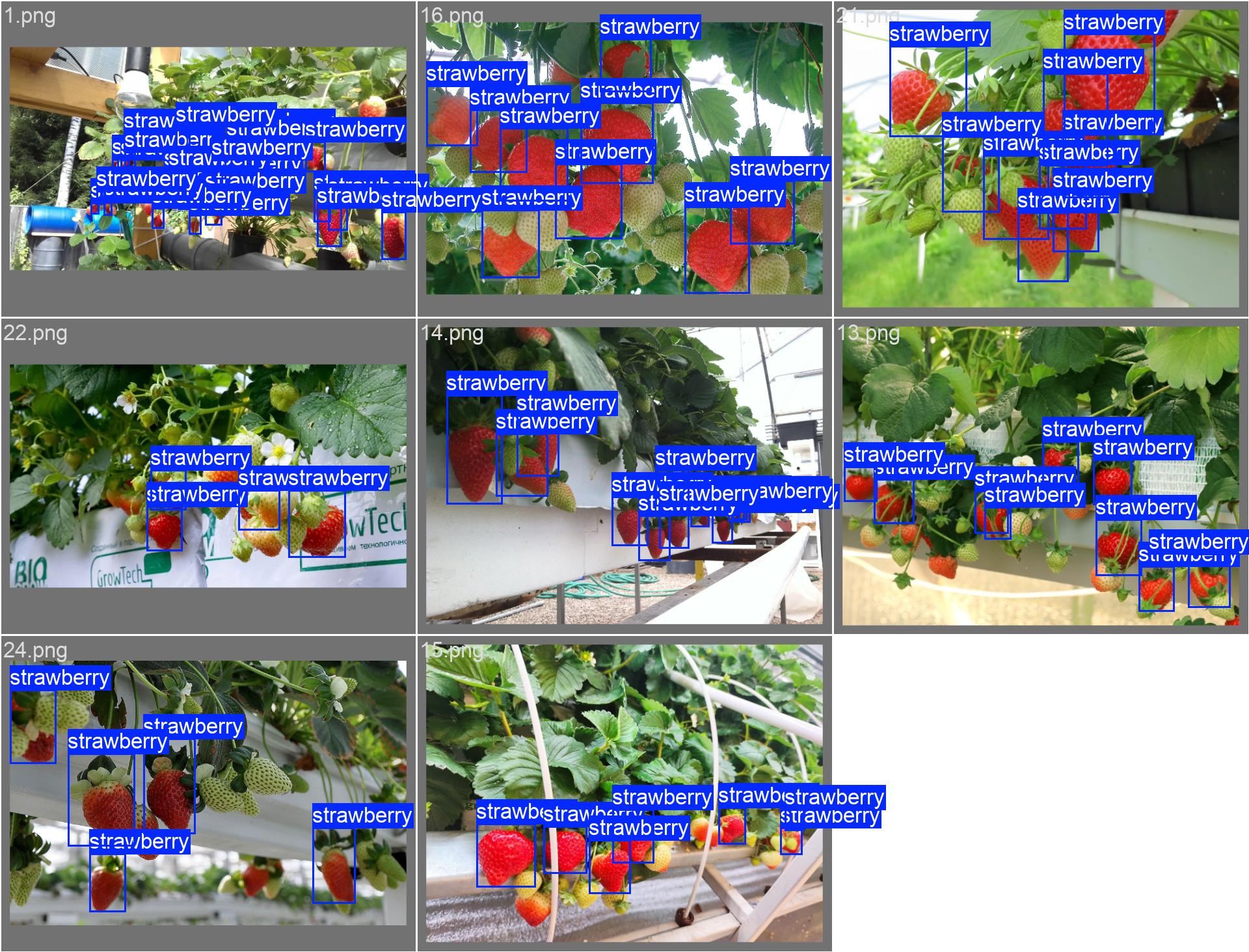} 
        \caption{Ground-truth labels}
        \label{fig:strawberry_label}
    \end{subfigure}
    \hfill
    \begin{subfigure}[b]{0.48\textwidth}
        \centering
        \includegraphics[width=\textwidth]{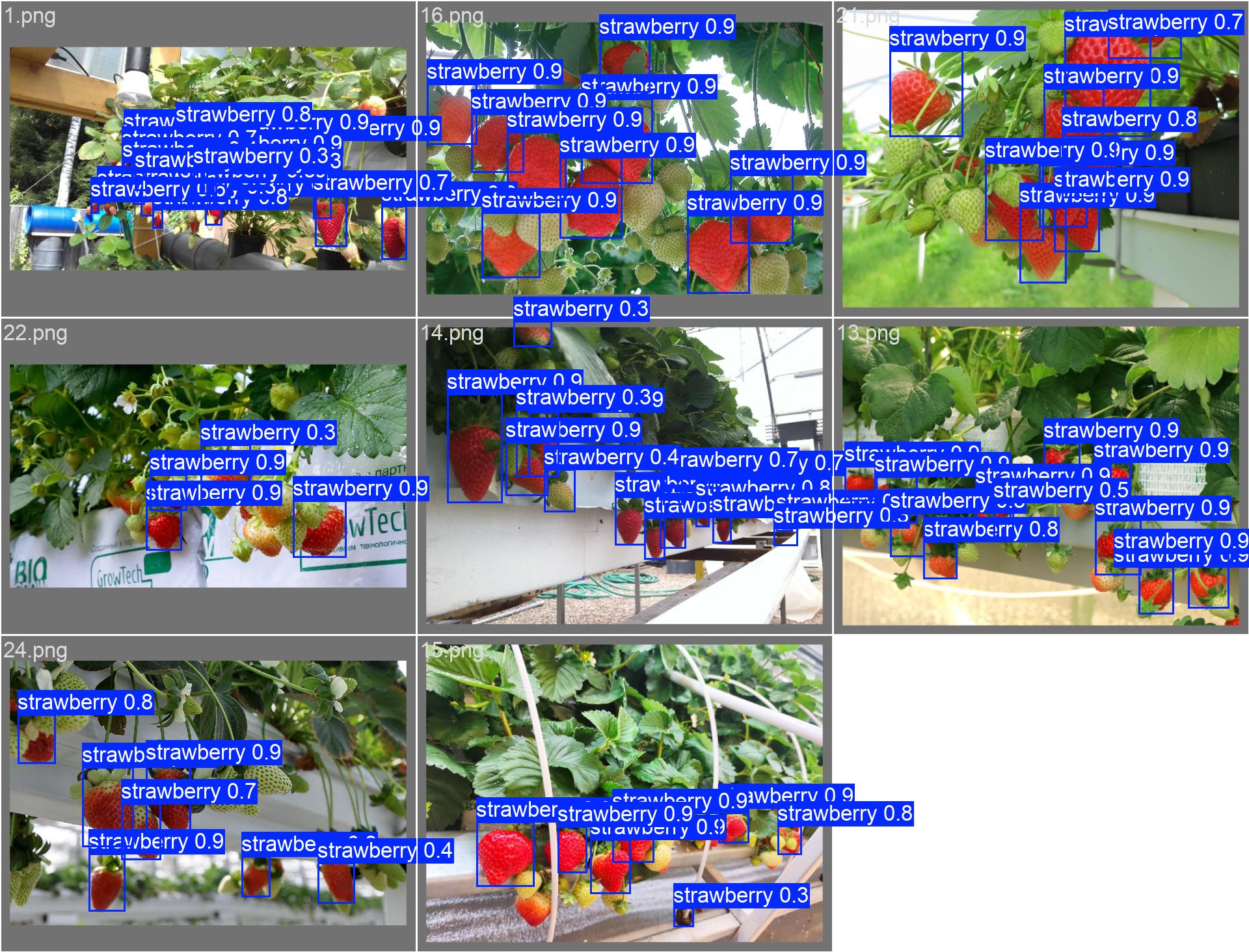} 
        \caption{Model predictions}
        \label{fig:strawberry_pred}
    \end{subfigure}
    \caption{Comparison of ground-truth labels and model predictions}
    \label{fig:strawberry_pred_overview}
\end{figure}

The model demonstrates strong detection capabilities for ripe red strawberries, as illustrated in Figure~\ref{fig:strawberry_pred_overview}. The left panels show the ground-truth annotations (labels), while the right panels present the model's predictions. High-confidence detections, particularly for ripe, red strawberries, consistently align well with the labels, confirming the model’s effectiveness in identifying mature fruits across different scales and dense greenhouse environments (e.g., overlapping leaves and complex backgrounds). Nevertheless, some detections exhibit lower confidence or slight misalignments, especially for partially occluded strawberries, unripe fruits, or those in challenging lighting conditions. Occasional false positives are observed, where bounding boxes are placed on non-strawberry objects or background elements. In dense clusters, overlapping bounding boxes may occur, suggesting that Non-Maximum Suppression (NMS) parameters could be further optimized. Overall, despite these minor limitations, the model shows robust detection performance across most scenarios.

\subsection{Robotic Arm Strawberry Search and Approach}
\subsubsection{Hand--Eye Calibration Results}

To quantitatively evaluate the performance of different hand--eye calibration methods, both translation and rotation errors are compared. The translation error is measured in meters, while the rotation error is measured in degrees.

\begin{table}[h]
\centering
\caption{Hand--Eye Calibration Results}
\begin{tabular}{lccccc}
\hline
Method & $t_x$ (m) & $t_y$ (m) & $t_z$ (m) & $\|t\|$ (m) & Rotation ($^\circ$) \\
\hline
Tsai      & -0.0132 &  0.0622 & -0.0739 & 0.0974 & 169.2 \\
Park      & -0.0178 &  0.0603 & -0.0774 & 0.0997 & 173.9 \\
Horaud    & -0.0178 &  0.0603 & -0.0774 & 0.0997 & 173.9 \\
Andreff   & -0.0263 &  0.0457 & -0.0223 & 0.0572 & 174.0 \\
Daniilidis& -0.0285 &  0.0450 & -0.0750 & 0.0920 & 174.6 \\
\hline
\end{tabular}
\end{table}

\begin{table}[h]
\centering
\caption{Overall Calibration Statistics}
\label{tab:calibration_statistics}
\begin{tabular}{lc}
\hline
Metric & Value \\
\hline
Number of samples & 45 \\
Translation consistency (std) & 0.0235 m \\
Best method & Park \\
Final translation magnitude & $\sim$0.10 m \\
Rotation & $\sim$174$^\circ$ \\
\hline
\end{tabular}
\end{table}

The experimental results demonstrate notable differences in performance among the evaluated hand--eye calibration methods. In terms of translation accuracy, the Andreff method achieves the lowest translation error of 0.0572 m, significantly outperforming the other methods, whose errors range from approximately 0.0920 m to 0.0997 m. This indicates that Andreff provides the most accurate estimation of the translational component.

Although the Andreff method reports the smallest translation magnitude (\( \|t\|=0.0572\ \mathrm{m} \)), method selection in this work is not based on a single minimum-error snapshot. For closed-loop robotic operation, calibration stability across pose pairs is more critical than occasional low translation estimates, because unstable extrinsics propagate directly to target projection and approach control.

To prioritize deployment reliability, this project used cross-sample consistency as the primary criterion, and absolute error as a secondary criterion. Under this criterion, the Park method yields the most stable behavior over 45 samples, with translation consistency (standard deviation) of \(0.0235\ \mathrm{m}\), while maintaining a physically plausible translation magnitude (about \(0.10\ \mathrm{m}\)) that matches the expected camera--gripper offset.

Rotation estimates from all methods are concentrated in a narrow range (\(169.2^\circ\) to \(174.6^\circ\)), indicating that rotational alignment is broadly consistent across solvers and that the dominant practical difference lies in translational robustness. In our pipeline, this robustness is more important because small translation fluctuations can produce visible target drift in base-frame localization and increase corrective motion during closed-loop approach.

Therefore, Park is selected as the final hand--eye solution because it provides the best trade-off between geometric plausibility and repeatability, leading to more stable end-to-end manipulation performance.

\subsection{Non-Destructive Sweetness Measurement}
\subsubsection{Classification with ToF}
Table~\ref{tab:grid_results} presents the performance (accuracy) of the proposed 3D CNN model for different voxel grid resolutions ($G = 32, 64, 128$) across 20 strawberry samples. 
The accuracy values indicate the model's prediction performance for each individual sample under varying spatial resolutions.

\begin{table}[H]
\centering
\caption{Performance comparison across different grid resolutions for each strawberry sample.}
\label{tab:grid_results}
\begin{tabular}{c|ccc}
\hline
\textbf{Sample} & \textbf{$G=32$} & \textbf{$G=64$} & \textbf{$G=128$} \\
\hline
1  & 0.9272 & 0.9423 & 0.9493 \\
2  & 0.8238 & 0.8170 & 0.8670 \\
3  & 0.8007 & 0.8840 & 0.8035 \\
4  & 0.9017 & 0.8685 & 0.9247 \\
5  & 0.8415 & 0.7937 & 0.7885 \\
6  & 0.9480 & 0.9467 & 0.9442 \\
7  & 0.8510 & 0.8658 & 0.8238 \\
8  & 0.8907 & 0.8935 & 0.9262 \\
9  & 0.8305 & 0.8363 & 0.8777 \\
10 & 0.8963 & 0.8885 & 0.9218 \\
11 & 0.9432 & 0.8852 & 0.9587 \\
12 & 0.9490 & 0.9327 & 0.9455 \\
13 & 0.8197 & 0.8790 & 0.8558 \\
14 & 0.9263 & 0.9362 & 0.9475 \\
15 & 0.8892 & 0.8385 & 0.9197 \\
16 & 0.8550 & 0.8720 & 0.8892 \\
17 & 0.9348 & 0.9448 & 0.9537 \\
18 & 0.8950 & 0.9218 & 0.9320 \\
19 & 0.9087 & 0.9085 & 0.9592 \\
20 & 0.8805 & 0.8862 & 0.9132 \\
\hline
\textbf{Mean} & \textbf{0.8899} & \textbf{0.8885} & \textbf{0.9092} \\
\hline
\end{tabular}
\end{table}

The results in Table~\ref{tab:grid_results} show that increasing the voxel grid resolution from $G=32$ to $G=128$ generally improves performance, with mean accuracy rising from $0.8899$ to $0.9092$ (an increase of approximately 2\%). This suggests that higher spatial resolution enables the 3D CNN to capture finer geometric and intensity variations within the strawberry point clouds.  

However, the improvement is not strictly monotonic across all samples. For example, in Sample~3, the accuracy at $G=64$ (0.8840) exceeds both $G=32$ (0.8007) and $G=128$ (0.8035), indicating that excessive resolution does not always guarantee better results. Possible explanations include:  
\begin{enumerate}
    \item \textbf{Overfitting:} Larger grids increase the number of input voxels, raising the risk of overfitting given the limited dataset size.  
    \item \textbf{Noise Sensitivity:} Higher resolution captures more fine-grained details but may also introduce high-frequency noise from raw point clouds, degrading prediction quality.  
    \item \textbf{Computational Complexity:} $G=128$ substantially increases memory usage and computation, potentially limiting batch size and affecting optimization stability.  
\end{enumerate}

In summary, although $G=128$ achieved the highest mean accuracy, the gain over $G=32$ and $G=64$ was limited (about 2\%), and not all samples benefited from the higher resolution. For applications prioritizing efficiency, $G=32$ may offer a more balanced trade-off between performance and computational cost.

\subsection{End-to-end results}
\subsubsection{Quantitative Results}
The proposed system achieved an end-to-end success rate of 88.10\% over 42 trials, 
demonstrating strong overall reliability in real-world operation. 
A stage-wise analysis reveals that the approach component (L3) achieved a perfect success rate of 100\% 
given successful detection, indicating that the motion planning and control pipeline is highly robust. 
This suggests that the incremental IK-based approach, combined with subgoal interpolation, 
is effective for stable and precise manipulation.

The detection stage (L1) achieved a high success rate of 95.24\%, 
with only a small number of failures (1 out of 42 trials). 
This indicates that the perception module is generally reliable, 
although it remains sensitive to certain edge cases such as challenging viewpoints or lighting conditions.

The primary source of failure lies in the sweetness estimation stage (L4), 
which achieved a conditional success rate of 92.50\%. 
Most failures were attributed to the inability to extract a valid sensing region 
(\texttt{sweetness\_no\_valid\_region}), suggesting that the limitation is not in the classifier itself, 
but in upstream sensing quality and region selection.

The remaining non-sweetness failures were observed in two specific categories: 
(1) temporary TF transform unavailability/timeout between perception and planning frames, and 
(2) IK/planning retry-limit exceedance when the target approached workspace feasibility boundaries. 
Although each occurred only once, these cases indicate that runtime transform robustness and boundary-aware motion fallback remain important for further reliability improvements.

Timing analysis shows a relatively large variance in completion time 
(mean 101.20s, max 236.33s), which is primarily due to repeated retries 
and occasional delays in perception-action synchronisation. 
This indicates that while the system is functionally reliable, 
there is room for optimisation in execution efficiency and pipeline coordination.

The RGB recentering (RC) mechanism showed limited usage (3 attempts), 
but a moderate success rate (66.67\%). 
Interestingly, sweetness estimation succeeded in all RC-triggered cases, 
suggesting that improved visual alignment can positively impact sensing quality. 
However, the small sample size limits the statistical significance of this observation.

\begin{table}[H]
  \centering
  \caption{End-to-end performance for the strawberry approach task.}
  \label{tab:e2e_results}
  \begin{tabular}{l c}
    \hline
    \textbf{Metric} & \textbf{Value} \\
    \hline
    Total trials $N$ & 42 \\
    Success count & 37 \\
    Success rate (\%) & 88.10\% \\
    \hline
    \multicolumn{2}{l}{\textit{Pipeline success rates}} \\
    Detection (L1) & 95.24\% (40/42) \\
    Approach (L3 $|$ L1) & 100.00\% (40/40) \\
    Sweetness (L4 $|$ L3) & 92.50\% (37/40) \\
    \hline
    \multicolumn{2}{l}{\textit{Failure breakdown (count / \%)}} \\
    No detection & 1 / 2.38\% \\
    Sweetness no valid region & 2 / 4.76\% \\
    TF transform unavailable/timeout & 1 / 2.38\% \\
    IK/planning retry limit exceeded & 1 / 2.38\% \\
    \hline
    \multicolumn{2}{l}{\textit{Timing statistics}} \\
    Mean completion time (s) & 101.20 \\
    Median completion time (s) & 107.10 \\
    Max completion time (s) & 236.33 \\
    \hline
    \multicolumn{2}{l}{\textit{RGB recenter (RC)}} \\
    RC success rate & 66.67\% (2/3) \\
    L4 success when RC=ok & 100.00\% (2/2) \\
    L4 success when RC=not-ok & 100.00\% (1/1) \\
    \hline
  \end{tabular}
\end{table}

Overall, the results indicate that the system is robust in control and reasonably reliable in perception, 
with the main bottleneck residing in sensing robustness and data quality for sweetness estimation. 
Future improvements should therefore focus on enhancing depth reliability, 
improving region selection strategies, and optimising system efficiency.

\subsubsection{Qualitative Results}
I qualitatively evaluate the proposed system in real-world scenarios to assess its robustness and behavior under varying conditions. 
The system is tested in an end-to-end manner, where perception, localization, and control are tightly integrated in a closed-loop pipeline.

\paragraph{Overall Behavior}
In most trials, the robotic arm is able to progressively approach the target strawberry in a smooth and stable manner. 
The system continuously updates the estimated target pose as new visual observations are received, resulting in gradual trajectory refinement. 
This closed-loop behavior allows the robot to compensate for perception noise and kinematic inaccuracies, avoiding large deviations from the target. 
In particular, the adaptive step-size strategy ensures that the motion remains feasible even when the initial target estimation is imperfect.

\begin{figure}[htbp]
    \centering
    \begin{subfigure}{0.32\linewidth}
        \centering
        \includegraphics[width=\linewidth]{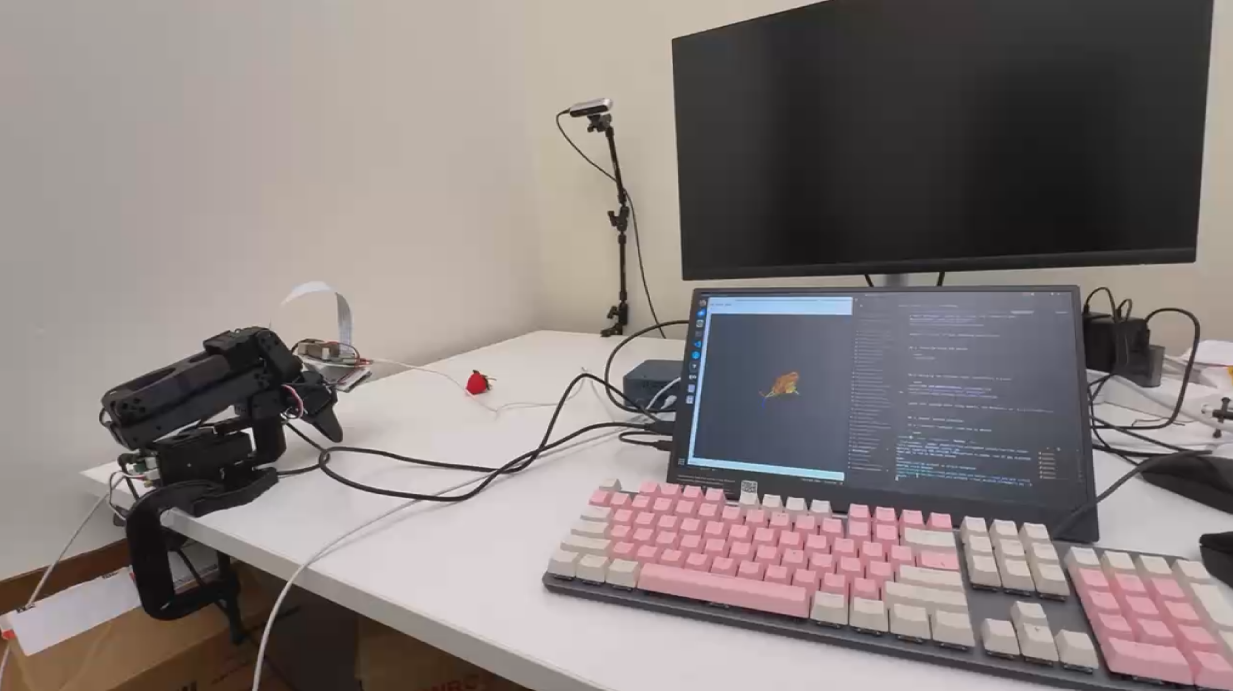}
        \caption{Initial detection}
    \end{subfigure}
    \hfill
    \begin{subfigure}{0.32\linewidth}
        \centering
        \includegraphics[width=\linewidth]{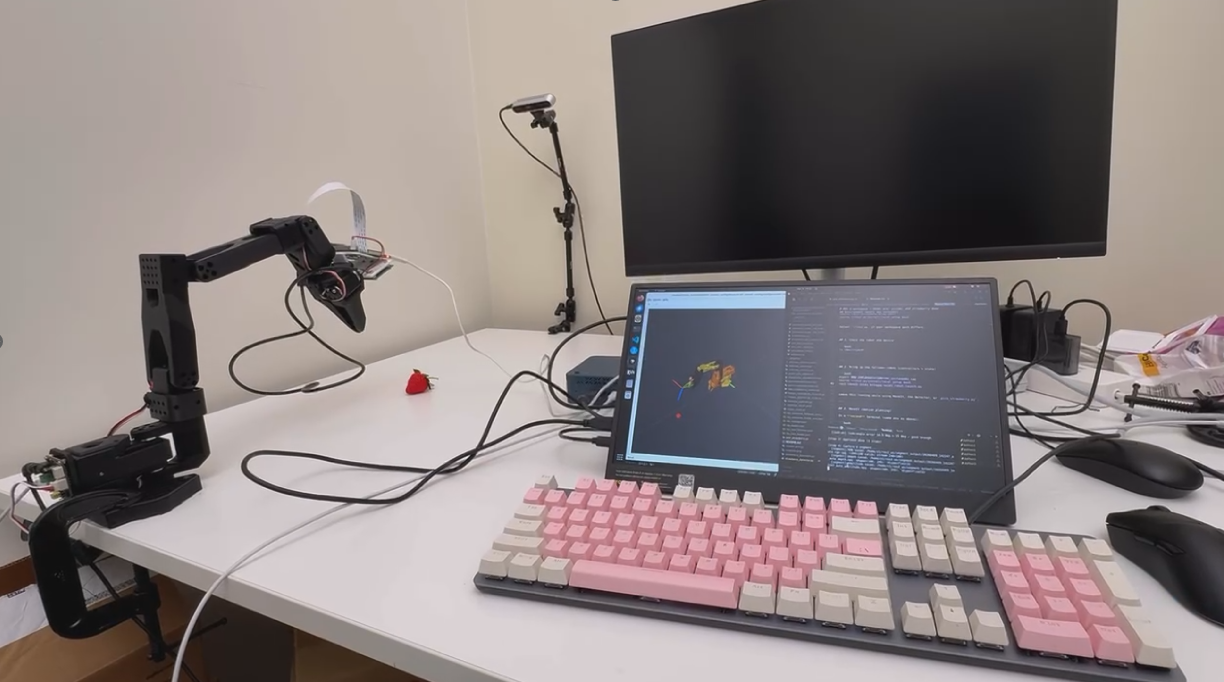}
        \caption{Mid approach}
    \end{subfigure}
    \hfill
    \begin{subfigure}{0.32\linewidth}
        \centering
        \includegraphics[width=\linewidth]{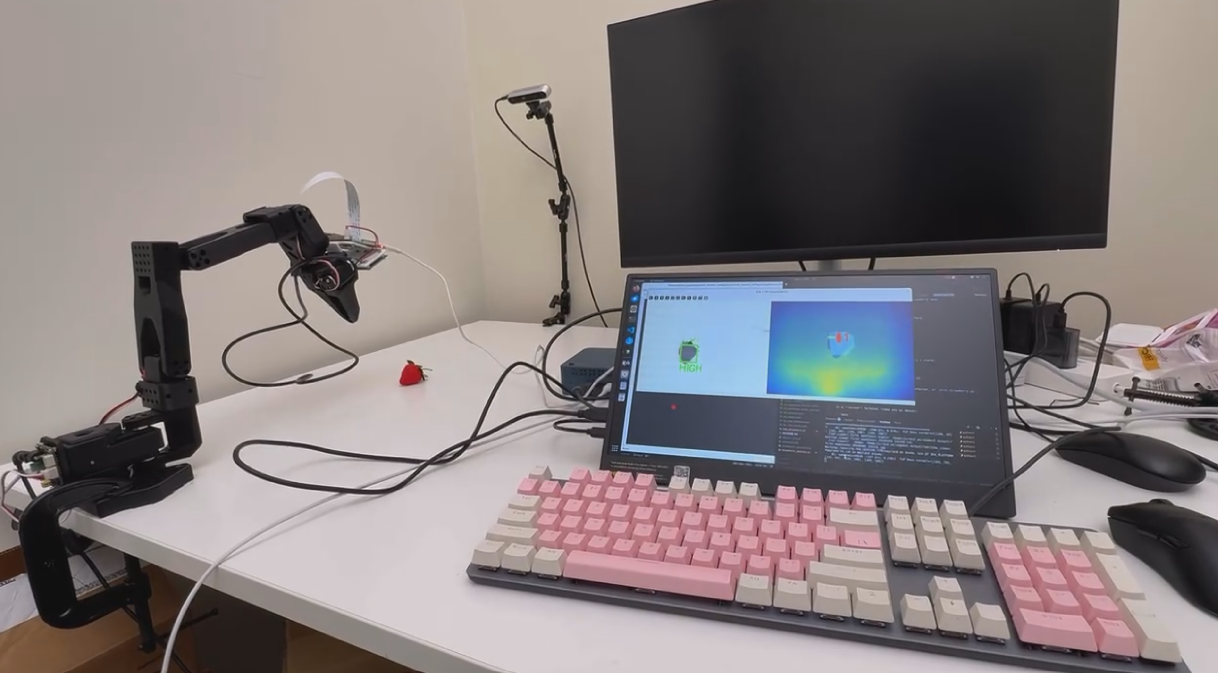}
        \caption{Final close-up}
    \end{subfigure}

    \caption{End-to-end approach sequence. The robot progressively refines its motion based on visual feedback from initial detection to final close-range positioning.}
    \label{fig:e2e_sequence}
\end{figure}

\paragraph{Representative Scenarios}
Several representative behaviors are observed during experiments. 
First, when the initial detection is slightly offset, the robot is able to correct its trajectory over multiple iterations, converging towards the true target position. 
Second, when the camera viewpoint changes significantly due to arm motion, the system successfully re-localizes the strawberry and adjusts the motion plan accordingly. 
Third, in cluttered environments with multiple red objects, the combined detection and segmentation pipeline helps maintain focus on the correct target, reducing false positives.

\paragraph{Failure Cases and Limitations}
Despite the overall robustness, several failure modes are observed. 
When the depth measurements are highly noisy or partially missing, the estimated target position may become unstable, leading to oscillatory motion. 
In addition, when the strawberry is partially occluded or located at the edge of the field of view, detection accuracy decreases, which may result in incorrect localization. 
Finally, extreme poses may lead to inverse kinematics failure, especially when the target lies near the boundary of the robot's reachable workspace. 
In such cases, the system attempts to reduce the step size, but may still fail if no feasible solution exists.

\section{Limitations and Future Improvements}
\subsection{Robotic Arm Strawberry Search and Approach}
\subsubsection{Waypoint-based scanning strategy}
While the waypoint-based scanning strategy is simple and efficient, it introduces several inherent limitations.

\paragraph{Insufficient viewpoint diversity leading to missed detections under occlusion.} 
The use of only two discrete viewpoints restricts the spatial coverage of the scene. Although the vertical scanning motion enables basic observation of upper and lower regions, it may still miss strawberries that are heavily occluded by leaves, stems, or other fruits. In such cases, additional viewpoints or more sophisticated exploration strategies would be required to improve visibility and detection robustness.

\paragraph{Lack of adaptability in fixed waypoint-based exploration.} 
The fixed waypoint configuration lacks adaptability, as the robot follows a predefined motion pattern regardless of the scene content. This can lead to inefficient exploration, especially when targets lie slightly outside the scanned region. The limitation becomes more pronounced in dynamic or unstructured environments where fruit positions are not predictable.

\paragraph{Suboptimal viewpoint selection for perception quality.} 
Although joint-space waypoints simplify control, they do not explicitly optimize for perception quality. The selected poses may not always provide the best viewing angle for detection or segmentation, particularly under challenging lighting conditions or complex backgrounds.

These limitations suggest that while the current waypoint strategy is suitable for controlled environments, more advanced approaches—such as adaptive viewpoint planning or learning-based exploration (e.g., VLA methods) could further improve performance and robustness in real-world scenarios.

\paragraph{Limited search region due to strong spatial assumptions.} 
The approach relies on the assumption that strawberries are located within a predefined frontal region of the robotic arm. In more realistic agricultural environments, such as dense greenhouse setups, fruits may appear outside this constrained field of view. As a result, the robot may fail to detect targets that lie beyond the predefined search area, limiting the generalizability of the system

\subsubsection{3D Localization}

The current 3D localization method relies on projecting a 2D reference point (centroid or bounding-box centre) into 3D space using depth measurements. 
This approach introduces several limitations.

\paragraph{Sensitivity to depth noise}
The estimated 3D position is sensitive to depth noise, especially when using ToF sensors under reflective surfaces or non-uniform lighting. 
Even with median-based filtering, depth outliers may still bias the final estimation.
\paragraph{Geometric approximation error}
The use of a single 2D reference point does not fully capture the true geometric centre of the strawberry in 3D space. 
As a result, the estimated position may deviate from the actual fruit centre, particularly when the object has irregular shape or is viewed from oblique angles.
\paragraph{Impact of partial occlusion}
Partial occlusion caused by leaves or neighbouring fruits can lead to incomplete segmentation, resulting in invalid or sparse depth samples within the region of interest. 
This directly degrades the robustness of depth association and increases localization error.
\paragraph{Assumption of local depth consistency}
The current method assumes locally consistent depth within the segmented region, which may not hold in cluttered greenhouse environments where background points are incorrectly included.

\subsubsection{VLA Probation}

\paragraph{Motivation for VLA Methods}
While the custom hand-eye calibration approach improves efficiency and system integration, it still relies on accurate geometric modelling of the robot and sensor setup. 
In practice, this introduces several limitations. The calibration process is sensitive to sensor noise, mechanical inaccuracies, and environmental changes. 
Even small errors in the estimated transformation can propagate through the pipeline, leading to noticeable positioning errors during robotic execution. 
As a result, additional mechanisms such as closed-loop perception are often required to compensate for these inaccuracies.

\paragraph{Limitations of geometry-based pipelines}
Traditional pipelines explicitly decompose the system into perception, calibration, and control modules. 
While this modular design improves interpretability, it also introduces error accumulation across stages. 
In particular, inaccuracies in calibration and 3D localisation can directly affect downstream motion planning, limiting overall system robustness in real-world environments.

\paragraph{VLA formulation}
To address these limitations, Vision-Language-Action (VLA) models are considered as a potential alternative. 
Unlike traditional approaches, VLA methods aim to learn a direct mapping from visual observations to robot actions. 
In this paradigm, the system takes raw sensor input without explicitly estimating intermediate representations such as 3D coordinates or transformation matrices.

\paragraph{Preliminary experiments and observations}
A preliminary evaluation of a VLA-based policy was conducted over 50 episodes in the strawberry approach task. 
The observed success rate was approximately 40\%, which is significantly lower than the performance achieved by the structured ROS2-based pipeline.

This relatively low success rate can be attributed to several factors. 
First, the amount of training data was limited, which restricts the model's ability to generalise across different viewpoints, lighting conditions, and occlusion scenarios. 
Second, the learned policy exhibited instability in fine-grained control, particularly during the final approach phase where precise positioning is required. 
Third, the lack of explicit geometric constraints makes it difficult for the model to maintain consistent spatial reasoning, especially in cluttered environments.

\paragraph{Advantages of VLA methods}
Despite the current limitations, VLA-based approaches offer several promising advantages. 
They do not require explicit hand-eye calibration; instead, the mapping between perception and action is learned implicitly from data. 
This reduces dependence on precise geometric modelling and mitigates the accumulation of systematic calibration errors.

Furthermore, VLA models have the potential to adapt more effectively to unstructured environments. 
By learning directly from data, they can better handle variations in lighting, occlusion, and object appearance compared to rule-based pipelines.

\paragraph{Challenges of VLA methods}
However, these benefits come at the cost of increased data requirements and reduced interpretability. 
Training a robust VLA model typically requires a large-scale dataset of paired observations and actions, which is expensive to collect. 
In addition, the learned policy operates as a black box, making it difficult to diagnose failures or ensure safety in real-world deployment.

\paragraph{Design choice in this work}
Given these considerations, the traditional ROS2-based pipeline is adopted as the primary solution in this project due to its reliability, interpretability, and low data requirements. 
The experimental results suggest that, under limited data conditions, structured geometric pipelines remain more effective for precise manipulation tasks.

Nevertheless, VLA methods represent a promising future direction. 
With larger datasets, improved model architectures, and better integration of geometric priors, they may enable more flexible and scalable robotic perception and manipulation systems.

\subsection{Strawberry Detection and Segmentation}
\subsubsection{Robustness to Lighting Conditions}

The use of HSV colour space improves robustness to illumination changes compared with raw RGB thresholding, as hue is less sensitive to lighting intensity variations. However, extreme lighting conditions (e.g., strong shadows, specular reflections, or low saturation) can still degrade segmentation quality.

To mitigate this, the detection stage relies on the YOLO model as a primary cue, ensuring that candidate regions are first identified based on learned features rather than colour alone. The HSV-based segmentation is then applied locally within each detected bounding box, reducing the likelihood of false positives caused by background regions.

In addition, a minimum saturation constraint is applied within candidate regions to suppress weakly coloured areas. This helps reject background pixels under low-light conditions and improves the consistency of centroid estimation.

\subsubsection{Multiple Strawberry Handling}

The pipeline naturally supports multiple strawberries through the use of multi-instance detection. The YOLO detector returns a set of bounding boxes, each corresponding to a candidate strawberry instance.

For each detected bounding box, segmentation and centroid extraction are performed independently, resulting in a set of 2D reference points:
\[
\{(c_x^{(i)}, c_y^{(i)})\}_{i=1}^{M}
\]
where $M$ is the number of detected strawberries.

In the current implementation, a simple selection strategy is used to choose a target for manipulation. The system prioritises the strawberry that is closest to the camera, estimated based on the associated depth measurements. This heuristic favours reachable and less occluded targets.

More advanced strategies, such as ranking based on size, visibility, or predicted ripeness, could be incorporated in future work to improve task efficiency and decision-making.

\subsection{Non-Destructive Sweetness Measurement}

While the current system demonstrates promising performance, there remain several areas for enhancement to fully realise its potential.

\paragraph{Accuracy improvement}
Achieving a target RMSE of $<0.30\,^{\circ}\text{Brix}$ will require more advanced sensing and modelling strategies. 
This includes geometry-aware compensation leveraging ToF-derived surface normals, the adoption of multi-point sampling, and extending spectral coverage into the SWIR band to better capture sweetness-related features.

\paragraph{Continuous learning at the edge}
Incorporating dataset versioning and domain adaptation techniques would enable continuous improvement of the model. 
This allows the system to progressively enhance prediction accuracy by leveraging growing datasets and adapting to new cultivars, seasonal variations, and device conditions.

\paragraph{Scalability and robustness}
Further optimisation is required to maintain real-time performance while improving stability. 
In particular, the system should sustain a processing time of under 2 seconds per berry while achieving robust predictions across varying lighting conditions and at sweetness extremes.

\paragraph{Challenges in multi-modal fusion}
Compared to larger fruits such as apples, strawberries are significantly smaller, which increases the difficulty of implementing effective ToF and NIR data fusion. 
Reliable fusion requires a substantially larger dataset and careful model design to align spatial and spectral information.

Due to project time constraints, multi-modal fusion was not implemented in this work. 
However, it represents a key direction for future development and is expected to significantly improve prediction performance.

\section*{Acknowledgements}

I sincerely thank Professor Wen Hu for his invaluable guidance and mentorship throughout this project. His clear research direction and constructive feedback provided a strong foundation for the development of the system and helped shape the overall approach of this work.

I am also grateful to Dr. Mark Cardamis for his continuous support and insightful suggestions. In particular, his validation of the feasibility of non-destructive fruit sweetness measurement played a critical role in guiding the design of the sensing methodology. His input, from the initial hardware setup to subsequent model refinement, significantly contributed to the progress of this project.

I would like to express my sincere appreciation to Yu Hao for his extensive technical support. His assistance in addressing practical implementation challenges and facilitating the real-world deployment of the system was invaluable to the successful completion of this project.

I also extend my thanks to my peers and colleagues for their valuable support throughout this project. In particular, I acknowledge other members of Professor Wen Hu's research group for their helpful discussions and feedback. I would also like to thank my teammates for their collaboration on the implementation of the strawberry sweetness evaluation component.

Special thanks to Eason Lyu for his support in 3D modelling, which contributed to the design and implementation of the hardware components. I am also grateful to the Mechanical Engineering students who assisted with soldering the CSI connections, as well as the Makerspace team for providing 3D printing services that enabled rapid prototyping of the sensor mount.

\printbibliography
\appendix
\section{System Parameter Summary}
\label{app:params}

This appendix summarizes the key runtime and training parameters used in the integrated strawberry perception--manipulation--sweetness pipeline.
Unless otherwise stated, values are default settings from the current implementation.
Some parameters can be overridden via environment variables at runtime.

\subsection{Perception and Segmentation Parameters}
\label{app:params_perception}

\begin{table}[H]
\centering
\caption{RGB/HSV/YOLO perception parameters (default values).}
\label{tab:app_perception_params}
\small
\begin{tabular}{p{4.2cm}p{3.2cm}p{7.0cm}}
\hline
\textbf{Parameter} & \textbf{Default} & \textbf{Role} \\
\hline
\texttt{V2\_CONF\_THRESHOLD} & 0.2 & Detection confidence threshold (candidate acceptance). \\
\texttt{V2\_NMS\_THRESHOLD} & 0.34 & NMS IoU threshold for suppressing overlapping boxes. \\
\texttt{V2\_MAX\_DETECTIONS} & 1 & Maximum retained target instances for downstream control. \\
\hline
\texttt{HSV\_LOW\_1 / HIGH\_1} & [0,120,40] / [7,255,230] & Red range (lower hue interval) for HSV segmentation. \\
\texttt{HSV\_LOW\_2 / HIGH\_2} & [172,120,40] / [180,255,230] & Red range (upper hue interval) for hue wrap-around. \\
\texttt{MIN\_SOLIDITY} & 0.85 & Contour solidity filter for berry-like regions. \\
\texttt{ASPECT\_MIN / MAX} & 0.5 / 2.0 & Aspect-ratio gate for contour filtering. \\
\texttt{MIN\_EXTENT} & 0.60 & Region extent threshold for contour quality control. \\
\texttt{MIN\_CONTOUR\_AREA} & scaled, min 300 px & Reject very small segmented regions. \\
\hline
\texttt{MAX\_STEREO\_MATCH\_PX} & 25.0 px & Maximum 2D projection distance for RGB--ToF correspondence. \\
\hline
\texttt{V2\_HOMOGRAPHY\_DST\_DX\_PX} & 15 px & ToF homography offset correction in \(u\). \\
\texttt{V2\_HOMOGRAPHY\_DST\_DY\_PX} & 8 px & ToF homography offset correction in \(v\). \\
\hline
\end{tabular}
\end{table}

\subsection{Depth Linking and 3D Localization Parameters}
\label{app:params_depth_localization}

\begin{table}[H]
\centering
\caption{Depth sampling, robust filtering, and temporal stabilization parameters.}
\label{tab:app_depth_params}
\small
\begin{tabular}{p{4.2cm}p{3.2cm}p{7.0cm}}
\hline
\textbf{Parameter} & \textbf{Default} & \textbf{Role} \\
\hline
\texttt{FG\_CLUSTER\_LOW\_PCT} & 5.0\% & Remove nearest outlier fraction before foreground anchoring. \\
\texttt{FG\_WIN\_MM} & 120.0 mm & Foreground depth window \([Z_{\min}, Z_{\min}+120]\). \\
\texttt{FG\_MIN\_SAMPLES} & 8 & Minimum support points for foreground-cluster estimate. \\
\texttt{FG\_MIN\_SAMPLES\_STRICT} & 5 & Strict minimum before adaptive relaxation. \\
\hline
\texttt{DEPTH\_PATCH\_RADIUS} & 7 & Fallback local patch radius (\(15\times15\) window). \\
\texttt{DEPTH\_PATCH\_MIN\_VALID} & 5 & Minimum valid pixels in patch after masking. \\
\texttt{DEPTH\_TRIM\_PCT\_LOW/HIGH} & 12.0 / 88.0 & Trimmed statistics range for robust depth estimation. \\
\texttt{DEPTH\_TRIM\_MIN\_SAMPLES} & 9 & Minimum retained samples after trimming. \\
\hline
\texttt{AMP\_DEPTH\_GATE\_FRAC} & 0.06 & Amplitude-informed confidence gate for depth samples. \\
\texttt{AMP\_REGION\_MAX\_DISTANCE} & 30 px & Maximum region-growing distance in amplitude map. \\
\texttt{AMP\_DEPTH\_WEIGHT} & 0.05 & Depth-difference weight in amplitude-region growth. \\
\texttt{AMP\_MAX\_DEPTH\_CHANGE\_MM} & 100.0 mm & Neighbor depth-change cap in amplitude region growing. \\
\texttt{AMP\_INTENSITY\_FLOOR\_DIV} & 3.8 & Low-amplitude rejection scaling divisor. \\
\texttt{MIN\_AMP\_MASK\_PIXELS} & 80 & Reject tiny amplitude masks. \\
\hline
\texttt{MIN\_DEPTH\_MM / MAX\_DEPTH\_MM} & 30 / 1000 mm & Hard validity range for depth sampling. \\
\texttt{WORK\_Z\_CAM\_MIN\_M / MAX\_M} & 0.08 / 0.95 m & Valid camera-frame working depth range. \\
\texttt{DEPTH\_TEMPORAL\_WIN} & 5 & Sliding-median temporal stabilization window. \\
\texttt{CENTROID\_JUMP\_RESET\_PX} & 36 px & Reinitialize tracking when centroid jump is too large. \\
\hline
\end{tabular}
\end{table}

\subsection{Motion Planning and Closed-loop Control Parameters}
\label{app:params_control}

\begin{table}[H]
\centering
\caption{Approach control, IK optimization, and closed-loop execution parameters.}
\label{tab:app_control_params}
\small
\begin{tabular}{p{4.2cm}p{3.2cm}p{7.0cm}}
\hline
\textbf{Parameter} & \textbf{Default} & \textbf{Role} \\
\hline
\texttt{SEARCH\_WAYPOINTS} & 2 joint-space poses & Up/down scanning for strawberry search. \\
\texttt{MAX\_CLOSED\_LOOP\_STEPS} & 12 & Maximum refinement iterations during approach. \\
\texttt{STEP\_FRACTION} & 0.20 & Interpolation fraction per incremental approach step. \\
\hline
\texttt{CLOSE\_ENOUGH\_M} & 0.20 m & Stop criterion for approach proximity. \\
\texttt{LOOK\_DEG\_THRESHOLD} & 25.0 deg & Guard: do not stop if look-at error is too high. \\
\texttt{LOOK\_AT\_OK\_DEG} & 6.0 deg & Good alignment threshold (runtime tunable). \\
\hline
\texttt{CAPTURE\_CAM\_DIST\_M} & 0.18 m & Preferred camera-to-target capture distance. \\
\texttt{CAPTURE\_CAM\_DIST\_MIN} & 0.13 m & Minimum acceptable camera-to-target distance. \\
\hline
\texttt{REDETECT\_STABLE} & 3 & Required stable detections before acceptance. \\
\texttt{REDETECT\_TIMEOUT} & 4.0 s & Timeout for re-detection stage. \\
\texttt{REDETECT\_JUMP\_M} & 0.03 m & Reject abrupt target jumps beyond this threshold. \\
\hline
\texttt{REDETECT\_TRUST\_K} & 0.25 & Re-detection trust scaling factor. \\
\texttt{REDETECT\_TRUST\_BASE} & 0.05 & Base trust level for measurement fusion. \\
\texttt{REDETECT\_HARD\_CAP} & 0.60 & Hard cap for trust-bound update magnitude. \\
\texttt{REDETECT\_ALPHA\_SOFT} & 0.40 & Soft blending factor in target update. \\
\hline
\texttt{IK\_POS\_TOL} & 0.010 m & Accept IK only if Cartesian error \(<10\) mm. \\
\texttt{Z\_FLOOR} & 0.05 m & Minimum gripper height constraint in \texttt{base\_link}. \\
\texttt{Z\_PENALTY\_WEIGHT} & 1000.0 & Penalty weight for violating floor-height constraint. \\
\texttt{JOINT\_REG\_WEIGHT} & 0.0001 & Regularization weight for smooth joint updates. \\
\texttt{LOOK\_AT\_WEIGHT} & 0.0003 & Camera look-at alignment term weight in IK objective. \\
\hline
\texttt{REACH\_MAX\_M} & 1.50 m & Maximum reach bound for target feasibility checks. \\
\texttt{BACKOFF\_DIST\_M} & 0.08 m & Recovery backoff distance after repeated failures. \\
\texttt{HARDCAP\_BACKOFF\_M} & 0.10 m & Additional hard-cap backoff distance. \\
\texttt{HARDCAP\_BACKOFF\_LIMIT} & 2 & Maximum hard-cap backoff retries. \\
\hline
\end{tabular}
\end{table}

\subsection{Sweetness Model Parameters (ToF 3D-CNN)}
\label{app:params_sweetness}

\begin{table}[H]
\centering
\caption{ToF preprocessing and 3D-CNN parameters for sweetness classification.}
\label{tab:app_sweetness_params}
\small
\begin{tabular}{p{4.2cm}p{3.2cm}p{7.0cm}}
\hline
\textbf{Parameter} & \textbf{Default} & \textbf{Role} \\
\hline
Sweetness class threshold & \(8.5^\circ\)Brix & Binary split between low/high sweetness classes. \\
\texttt{GRID\_SIZE} & 32 & Voxel grid resolution for 3D-CNN input. \\
\texttt{target\_N} & 3841 points & Point-cloud padding/truncation length. \\
\hline
\texttt{max\_depth\_threshold} & 300 mm & Upper depth bound in ToF preprocessing. \\
\texttt{min\_depth\_threshold} & 130 mm & Lower depth bound in ToF preprocessing. \\
\texttt{min\_intensity\_threshold} & 2000 & Intensity gate for valid ToF points. \\
\texttt{max\_distance} & 35 px & Radius for connected region growing in ToF map. \\
\texttt{depth\_weight} & 0.5 & Depth penalty weight in region-growing rule. \\
\texttt{max\_depth\_change} & 50 mm & Max neighbor depth change in region expansion. \\
\hline
3D-CNN conv channels & 2\(\rightarrow\)8\(\rightarrow\)16\(\rightarrow\)32\(\rightarrow\)64 & Backbone channel configuration. \\
Pooling strategy & \(2\times2\times2\) max pooling (3 stages) & Progressive downsampling. \\
Readout & Global average pooling + FC(64\(\rightarrow\)256\(\rightarrow\)2) & Binary logits output. \\
Dropout schedule & Adaptive linear decay & Stronger regularization early, weaker near convergence. \\
\hline
\end{tabular}
\end{table}

\subsection{Experiment Runtime and Logging Parameters}
\label{app:params_runtime}

\begin{table}[H]
\centering
\caption{Runtime orchestration and logging parameters.}
\label{tab:app_runtime_params}
\small
\begin{tabular}{p{4.2cm}p{3.2cm}p{7.0cm}}
\hline
\textbf{Parameter} & \textbf{Default} & \textbf{Role} \\
\hline
\texttt{RETURN\_START\_SETTLE\_S} & 2.5 s & Post-return waiting time for stabilization. \\
\texttt{POST\_SWEETNESS\_HOLD\_S} & 60.0 s & Hold duration after sweetness capture stage. \\
\hline
\texttt{RN\_VALIDATE\_ENABLED} & true & Enable RetinaNet cross-check in re-detection stage. \\
\texttt{RN\_VALIDATE\_MARGIN\_PX} & 15 px & Margin for RetinaNet validation overlap. \\
\texttt{RN\_VALIDATE\_CONF} & 0.25 & RetinaNet validation confidence threshold. \\

\hline
\end{tabular}
\end{table}

\end{document}